\documentclass[journal]{IEEEtran}
\usepackage{amsmath,amssymb,amsfonts}
\usepackage{algorithm}
\usepackage{algpseudocode}
\usepackage{array}
\usepackage[caption=false,font=scriptsize,labelfont=scriptsize]{subfig}

\usepackage{textcomp}
\usepackage{stfloats}
\usepackage{url}
\usepackage{verbatim}
\usepackage{graphicx}
\usepackage{adjustbox}
\usepackage{cite}
\usepackage{xcolor}
\usepackage[colorinlistoftodos]{todonotes}

\begin{document}

\title{A General Control Method for \par Human-Robot Integration}

\author{Maddalena Feder$^{1,2}$, Giorgio Grioli$^{1,2}$, Manuel G. Catalano$^{1}$ and Antonio Bicchi$^{1,2}$ 

\thanks{$^{1}$Soft Robotics for Human Cooperation and Rehabilitation, Istituto Italiano di Tecnologia, Genova 16163, Italy
        ({\tt\small maddalena.feder@iit.it}).}
\thanks{$^{2}$Department of Information Engineering and Centro di Ricerca ``Enrico Piaggio", University of Pisa, 56122, Pisa, Italy.}}

\maketitle

\begin{abstract}
    This paper introduces a new generalized control method designed for multi-degrees-of-freedom devices to help people with limited motion capabilities in their daily activities. The challenge lies in finding the most adapted strategy for the control interface to effectively map user's motions in a low-dimensional space to complex robotic assistive devices, such as prostheses, supernumerary limbs, up to remote robotic avatars. The goal is a system which integrates the human and the robotic parts into a unique system, moving so as to reach the targets decided by the human while autonomously reducing the user's effort and discomfort.
    We present a framework to control general multi DoFs assistive systems, which translates user-performed compensatory motions into the necessary robot commands for reaching targets while canceling or reducing compensation.  The framework extends to prostheses of any number of DoF up to full robotic avatars, regarded here as a sort of ``whole-body prosthesis'' of the person who sees the robot as an artificial extension of their own body without a physical link but with a sensory-motor integration.  We have validated and applied this control strategy through tests encompassing simulated scenarios and real-world trials involving a virtual twin of the robotic parts (prosthesis and robot) and a physical humanoid avatar.
\end{abstract}


\section{Introduction} \label{sec:introduction}
\IEEEPARstart{A}{ssistive} and rehabilitation devices such as powered wheelchairs, assistive robotic arms, and limb prostheses play a crucial role in assisting individuals with severe motor impairments \cite{Argall:2017}, which require daily assistance due to e.g. spinal cord or brain injuries.

The chosen interfaces, such as joysticks, head arrays, and sip-and-puff, often pose challenges in managing assistive devices \cite{Argall:2015}. They have an impact on aspects such as timing, accuracy, and transient noise \cite{Argall:2020}.
These interfaces are low dimensional and operate discretely, necessitating users to switch between different areas of the control space. Partially autonomous control can reduce cognitive and physical burden on users \cite{Argall:2017}.

\begin{figure}[t]
	\centering
	{
		\includegraphics[width = \columnwidth]{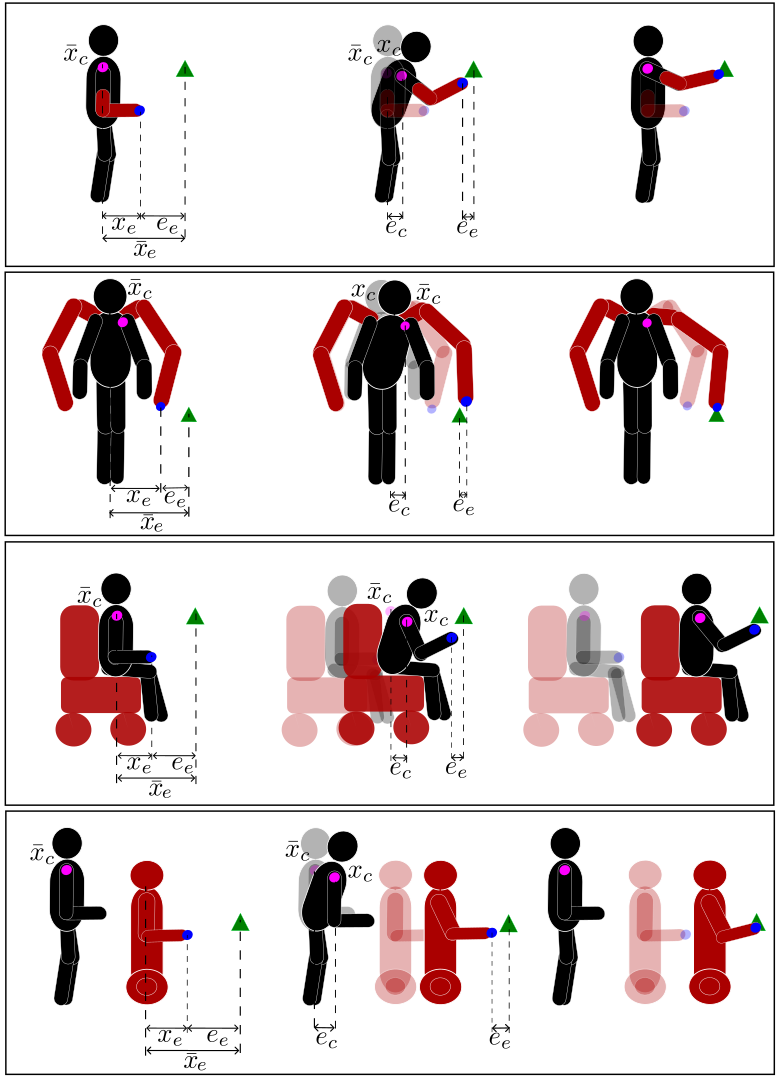} 
\caption{Starting, intermediate and final phases of the control
	strategy applied to different assistive robots. Human parts
	are depicted in black, while artificial parts are in
	red. Transparency is used to provide a reference to the
	starting conditions. The green triangle is the target to reach
	(known only to the human). The blue dot represents the
	end-effector frame, while the pink dot is the frame where we
	measure compensation. The rows illustrate applications of our
	framework to an upper limb prosthesis, to a supernumerary
	robot limb, to a wheelchair, and to a robot avatar. }
\label{fig:startModel}}
\end{figure}

In addition to techniques that exploit Inertial Measurement
Unit (IMU) sensors in \cite{Argall:2021} and brain-machine interfaces
(BMIs) in \cite{Hochberg:2012} to control assistive devices with high
number of Degrees of Freedom (DoFs), another approach involves
employing dimensionality reduction techniques, such as Principal
Component Analysis (PCA), \cite{Hochberg:2012}.  Beyond the branch of
assistive robotics, these control interfaces find applications also in
human-robot interfaces (HRIs), for both domestic and professional
settings.  Several studies implement different interfaces to control
robotic devices, such as upper-body movements \cite{Sanna:2013}, or
electromyographic sensors (EMG) on the forearm \cite{Stoica:2014}, or
sensorised upper-body soft exoskeleton, coupled with virtual reality
goggles, \cite{Stoica:2014}, or joint-level force control as the ones
for the supernumerary robotic limbs \cite{Prattichizzo:2021}. Another
interface system, which includes also a haptic admittance module, is
presented in \cite{Lamon:2020}.

In the prosthetic field, the use of EMG sensors is the most widely
adopted technique \cite{Farr:2018}. One of the main drawbacks is that
as the level of amputation rises, a greater number of sensors is
required to provide more inputs, while less independent measurement
sites are available. Additionally, a strategy becomes necessary to
switch the controller between multiple DoFs.  Furthermore, the user's
skin condition can impact control performance, as can changes in
electrode positions \cite{Roche:2014}. This control complexity may
lead prosthetic users to develop undesired compensatory motion habits,
which in turn, can result in postural issues and chronic pain \cite{Carey:2008}.

In recent years, researchers considered that compensatory motions of
the patient's body could be themselves useful to control prostheses.
\cite{Maimeri:2019} used IMUs on the stump in
combination with surface EMG signals to control multi-channel
prostheses and supernumerary limbs. \cite{Legrand:2020, Merad:2020, Legrand:2022} more explicitly
introduced the idea of exploiting body compensation as a signal to
control upper limb prostheses.
 
Inspired by this work, in this paper we generalize the idea of
using compensatory motions in the prosthesis control loop, by
including it in a more comprehensive dynamic system view which
encompasses both a model of the human user and of the prosthesis and
its controller, to be designed.

The generalization of the framework allows us to consistently include
widely varying systems, both in the human component, where different
abilities can be accommodated for, and in the artificial part, where
upper-limb prostheses can be extended to assistive devices and even
remote whole-body avatars. Preliminary results in this direction were
presented in \cite{Feder:2023}. In this paper, we substantially extend
and improve the framework introducing a model of the human reaching
and compensation control, and an observer which can predict the
reaching target intention and correct the controlled motion on--line.  

We apply this framework to three virtual prosthetic arms with
different numbers of joints. We also apply it to the control of a
robotic avatar. The avatar, Alter-Ego, is partly human--like, with a
torso, head, neck, two arms and hands, but locomotes on two wheels on
which it balances. We include the avatar in our control framework by
simply regarding it as a prosthesis of the whole body of the user. By
this approach, instead of being controlled through classical
teleoperation techniques mapping human motions to robot motions
one--to--one, the avatar becomes an artificial extension of the human
body. In this way, individuals with different motion capabilities are
enabled to command the robotic joints through their residual motions as
if it were their prostheses.

\section{Problem Definition} 
\label{sec:probDef}
Consider a dynamic system, composed of a person and a prosthesis of
any number of DoFs, up to an artificial replacement of the full body
as in an avatar. The system is modelled as a kinematic chain, with
$n_r$ robotic joints and $n_h$ human joints. Fig.~\ref{fig:startModel} provides a sketch representation of this model for both a prosthesis, a supernumerary limb, a wheelchair and an avatar.
We suppose that the user intent is specified in terms of reaching a
desired goal $\bar{x}_e$ with the end effector $x_e$, whose position
is a function of some or all the $n_h$ human joints $q_h$ and/or the
$n_r$ robotic joints $q_r$ as
\begin{equation}
    x_e = Q_e(q),
\end{equation}
where $q = \begin{bmatrix} q_h, q_r \end{bmatrix}^T$.  By defining the
human and robotic Jacobian matrices as $J_{he} = J_{he}(q)
= \frac{\partial Q_e}{\partial q_h}$ and $J_{re} = J_{re}(q)
= \frac{\partial Q_e}{\partial q_r}$, respectively, the differential
kinematics of the end effector is
\begin{equation} \label{eq:dxe_matrix}
\dot{x}_e = \begin{bmatrix} 
J_{he}(q) & J_{re}(q)
\end{bmatrix} 
\begin{bmatrix}
    \dot{q}_h \\ \dot{q_r}
\end{bmatrix}.
\end{equation}
We define the reaching error $e_e$ as
\begin{equation}
    e_e = \bar{x}_e - x_e.
\end{equation}
In the following, we assume that the reaching error $e_e$ is known to the user, based on the visual feedback of the displacement between the desired target and their hand. Notice also that both frames are expressed relative to the egocentric reference of the user, which is
usually placed in the head. An exception is the avatar case, where the egocentric reference is placed in the head of the avatar itself (through whose eyes the user sees the scene, cf. Fig.~\ref{fig:startModel}, last row).  

The task for us is to control $\dot{q}_r$ so that $e_e$ tends to zero, and to exploit a possible redundancy in the kinematics of the human-robot system to favor a relaxed configuration of the human joints $q_h$. The problem is not trivial because the desired final
position $\bar{x}_e$ is not available to the robot controller. To circumvent this problem, we exploit the ability of humans to include in their body schema devices and appendages \cite{MerleauPonty}, and to exploit them to reach their goals. For example, a person wearing even a completely passive prosthesis, such as a ``cosmetic hand'', can easily reach for an object, although at the cost of possibly exaggerated motions of the intact body joints in the shoulder, torso, waist and legs.

Consider hence the case that the user attempts to reach the goal without counting on the assistance of an active prosthesis, as they would do with a passive appendage of some form: the user will employ their own DoFs ($q_h$) while assuming that the prosthetic joints are
fixed ($\dot{q}_r = 0$) in the present configuration. The resulting body motions (which are somehow not natural and possibly fatiguing for the user) are commonly referred to as {\em compensatory movements}. The idea of exploiting compensatory motions in prosthesis
control was first introduced by \cite{Legrand:2022}. Let $x_c$ denote a compensatory reference frame fixed on a selected body segment (e.g. at the shoulder), let $\bar{x}_c$ indicate its posture in a relaxed body posture, and define a compensation error $e_c = \bar{x}_c -
x_c$.

In general, also the compensatory frame moves as a function of both the human and robotic joints as
\begin{equation}
    x_c = Q_c(q).
\end{equation}
Let $J_{hc} = J_{hc}(q) = \frac{\partial Q_c}{\partial q_h}$ and $J_{rc} = J_{rc}(q) = \frac{\partial Q_c}{\partial q_r}$ be the Jacobian matrices of the compensatory model with respect to the human and robot joints, respectively. We have 
\begin{equation} \label{eq:dxc_matrix}
    \dot{x}_c = \begin{bmatrix}
        J_{hc}(q) & J_{rc}(q)
    \end{bmatrix}
    \begin{bmatrix}
        \dot{q}_h \\ \dot{q}_r
    \end{bmatrix}.
\end{equation} 
In terms of reaching and compensation error dynamics, we write
\begin{equation} \label{eq:sys1}
    \begin{bmatrix}
        \dot{e}_e \\ \dot{e}_c
    \end{bmatrix} = 
    \begin{bmatrix}
        \dot{\bar{x}}_e \\ \dot{\bar{x}}_c
    \end{bmatrix} - 
    \begin{bmatrix}
        J_{he}(q) & J_{re}(q) \\ J_{hc}(q) & J_{rc}(q)
    \end{bmatrix} \begin{bmatrix}
        \dot{q}_h \\ \dot{q}_r
    \end{bmatrix}.
\end{equation}
Notice explicitly that the $J_{rc}$ term can be used to model cases where the robotic system influences the body compensation variables, as e.g. when an active exoskeleton is included.
Our goal in this paper is to design a controller for the robotic joints such that the prosthesis helps the user to reach their goal $x_e \rightarrow \bar{x}_e$, while allowing the compensatory motions to be reduced by bringing back $x_c \rightarrow \bar{x}_c$. In the
following, we will assume that the relaxed configuration $\bar{x}_c$ is constant.  In the error coordinates, the goal is simply to stabilize both $e_e$ and $e_c$ in zero. To design the controller, we will assume that the configurations of both robot and human joints are
measured in real time (via conventional encoders on the robot part, and motion capture systems for the human), so that $x_c$ and $x_e$ are known, and so is $e_c$. However, we cannot reasonably assume that $\bar{e}_e$ is known, as the human intended goal $\bar{x}_e$ is only
known to the human, and not available for designing the feedback controller.

\begin{figure}[t!] 
	\includegraphics[width = \columnwidth]{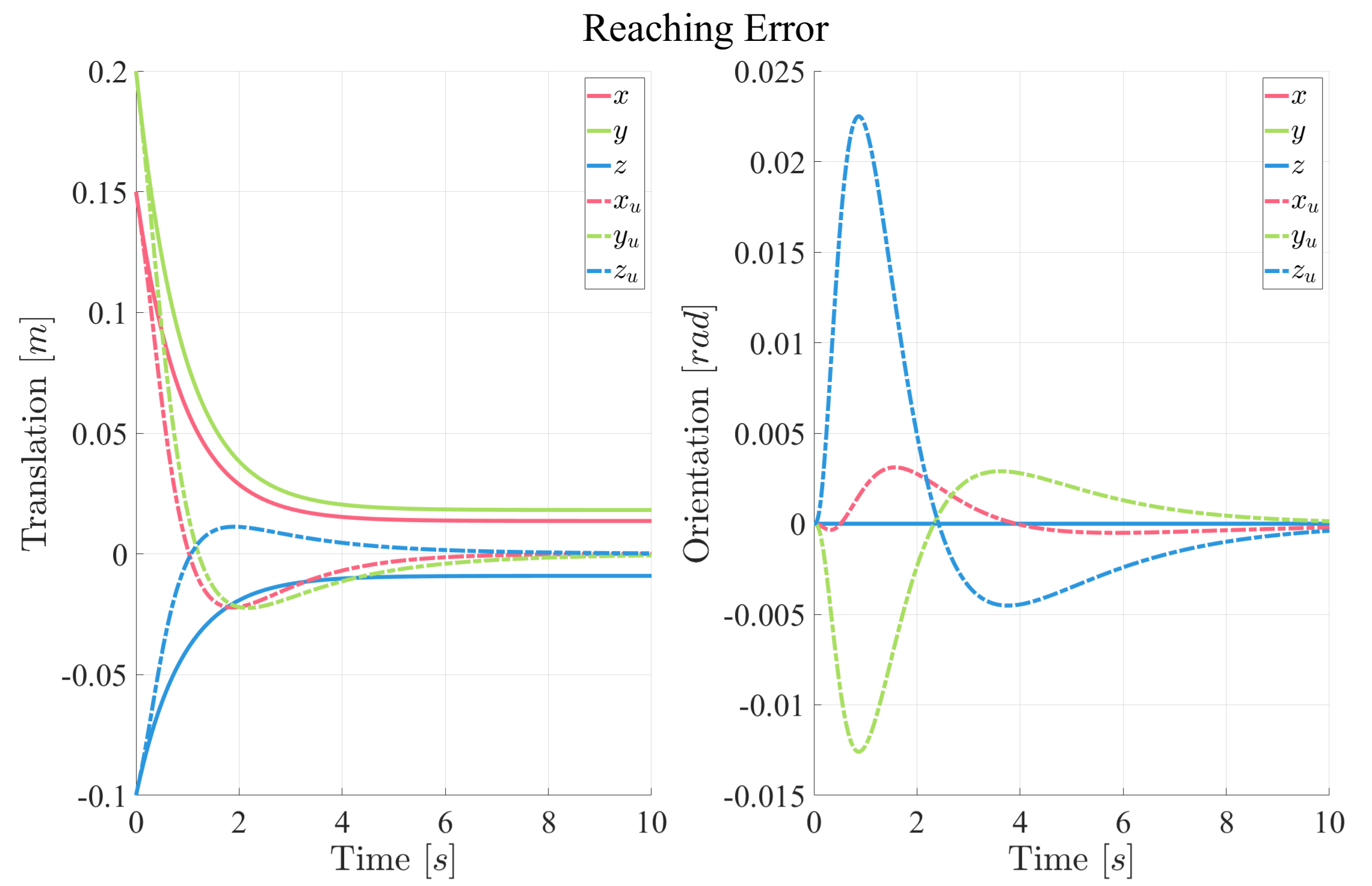} 
    \caption{Reaching error in a linear simulation: translation and orientation components of the reaching error $e_e$ without (continuous line) and with (dashed line) prosthetic controller. $\Lambda_e = 1\, I_6$ and $\Lambda_c = 0.1\, I_6$. The reaching error $e_e(0)$ is initialized at $\begin{bmatrix} 0.15, 0.2,-0.1 \end{bmatrix}$ for the translation component, and zero for the orientation. \label{fig:ReachingError}}
\end{figure}

\begin{figure}[t!] 
	\includegraphics[width = \columnwidth]{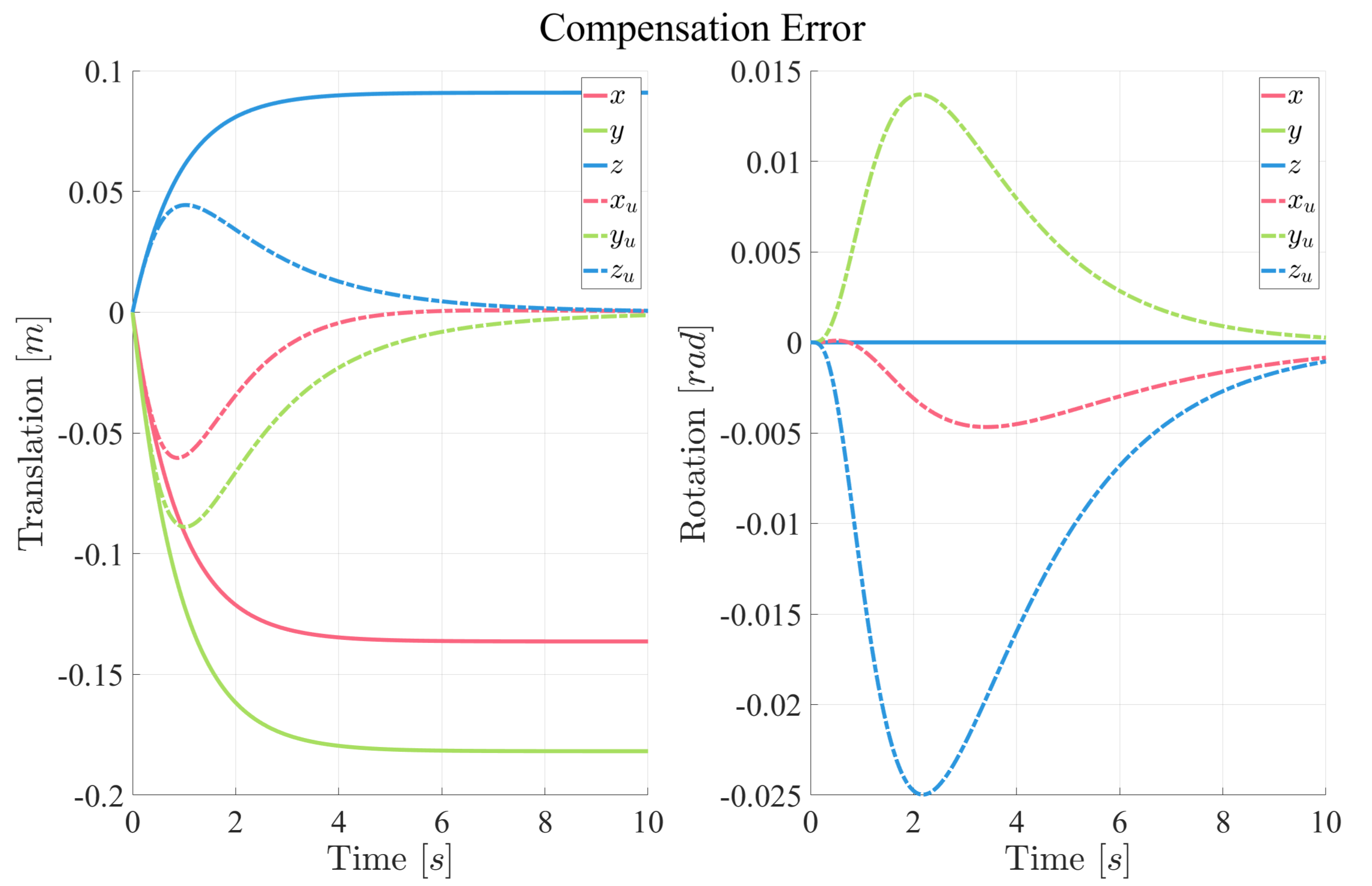} 
    \caption{Compensation error in a linear simulation: translation and orientation components of the compensation error $e_c$ without (continuous line) and with (dashed line) prosthetic controller. $\Lambda_e = 1\, I_6$ and $\Lambda_c = 0.1\, I_6$. Initial conditions on the compensation error $e_c(0)$ are set to zero. \label{fig:CompensationError}}
\end{figure}

\section{Approach} 
\label{sec:approach}
To illustrate the proposed solution to the problem presented in the
previous section, we start from a simplified analysis of the system in
proximity of the desired equilibrium $(e_e, e_c) = 0$.
To model the relationship between the human motion
and the reaching and compensation error, we assume a most simple
linear proportional model, i.e. we assume that the human tends to move
so as to reduce the reaching error target
\begin{equation} \label{eq:ass1}
    \widehat{J}_{he}\dot{q}_h = \Lambda _e e_e,
\end{equation}
while at the same time trying to reduce compensation through
\begin{equation} \label{eq:ass2}
    \widehat{J}_{hc}\dot{q}_h = \Lambda_c e_c,
\end{equation}
where $\Lambda_e$, $\Lambda_c$ are generic diagonal (or, more generally,
positive definite symmetric) square gain matrices.  The matrices
$\widehat{J}_{he}$ and $\widehat{J}_{hc}$ symbolically represent here
the human internal model of the kinematic maps, i.e. how the subject
expects her/his movements to be reflected in the hand and the
compensatory movements, respectively. We will discuss the realism of
these assumptions, and the choice of parameters, later.
Of course, it will not be possible for the human joint velocities
$\dot{q}_h$ to satisfy both equations at the same time, so the system of \eqref{eq:ass1} and \eqref{eq:ass2} is inconsistent in general. It can be assumed that a
solution for human joint velocities $\dot q_h$ is chosen by subjects
that weighs the reaching and compensation goals differently. This can
be formalized as an optimization problem with cost 
\begin{equation}
  \label{eq:qhopt}
C_w(\dot q_h) = w \| \widehat{J}_{he} \dot q_h - \Lambda_e e_e\|^2 + (1-w) \|
\widehat{J}_{hc} \dot q_h - \Lambda_c e_c\|^2
\end{equation}
for some $w \in [0, 1]$.  For notational simplicity, define
\[
\widehat{J}_h = \left[\begin{array}{c}
\widehat{J}_{he} \\ \widehat{J}_{hc}
  \end{array}\right],
J_r = \left[\begin{array}{c}
J_{re} \\J_{rc}
  \end{array}\right],
\Lambda = \left[\begin{array}{cc}
\Lambda_e & 0 \\ 0 & \Lambda_c
  \end{array}\right],
\]
and
\[
\xi = \left[\begin{array}{c}
  e_e \\ e_c
    \end{array}\right],
W =  \left[\begin{array}{cc} 
w I & 0 \\ 0 & (1-w) I
  \end{array}\right].
\]
In this more compact notation, (\ref{eq:ass1}) and (\ref{eq:ass2}) are
stacked as
\begin{equation}
\label{eq:Jstack}
\widehat{J}_{h} \dot{q}_h = \Lambda \xi
\end{equation}
while the index to be optimized in (\ref{eq:qhopt}) can be rewritten
as
\[
C_w(\dot q_h) = \left(\widehat{J}_h \dot q_h - \Lambda \xi\right)^T W \left(\widehat{J}_h \dot q_h - \Lambda \xi\right).
\]
It can be easily shown that, if the matrix $J_w := \widehat{J}_h^T W \widehat{J}_h$
is invertible, the optimal solution to (\ref{eq:Jstack}) is 
\begin{equation} \label{eq:dqh}
  \widehat{\dot{q}_h}  = J_w^{-1} \widehat{J}_{h}^T W \Lambda \xi.
\end{equation}
Notice that to have invertibility of $J_w$ it is sufficient that $\widehat{J}_h$
has full column rank, i.e. that all human joints $q_h$ considered in
the model participate in controlling either the task or compensation
frames. This can be assumed to happen, up to possibly removing
redundant joints from the human body model\footnote{Equivalently, the
solution can be expeditely obtained by solving
$\widehat{J}_h^T W \widehat{J}_h \dot q_h = \widehat{J}_h^T W \Lambda \xi $
as $\widehat{\dot{q}_h}  =J_w^{\dagger} \widehat{J}_{h}^T W \Lambda \xi.$}.

Substituting (\ref{eq:dqh}) in \eqref{eq:sys1} we get
\begin{equation} \label{eq:sys}
 \dot \xi = \dot {\bar x} - J_h J_w^{-1} \widehat{J}_{h}^T W \Lambda \xi
-  J_r \dot q_r.
\end{equation}

This equation can be regarded as a dynamic system with state $\xi$,
input $u=\dot q_r$, exogenous reference $r = \dot{\bar x}
= \begin{bmatrix} \dot{\bar x}_e \\ \dot{\bar x}_c \end{bmatrix}$, and
measurable output $y=e_c$, and rewritten as
\begin{equation} \label{eq:sysCanon}
    \begin{cases}
        \dot{\xi}(t) = A \xi(t) + B u(t) + r(t) \\
        y(t) = C \xi(t) + Du(t)
    \end{cases},
\end{equation}
where  
\begin{align}
\begin{split}
  A = - J_h J_w^{-1} \widehat{J}_{h}^T W \Lambda
\end{split}
\begin{split}
        B = -J_r,
\end{split}
\end{align} 
\begin{align}
\begin{split}
    C = \begin{bmatrix}
        0 & I_6
    \end{bmatrix}, \\
\end{split}
\begin{split}
    D = 0.
\end{split}
\end{align}
It should be noticed that, because the Jacobians are functions of the
configurations $q$ and so are the states, the system in
(\ref{eq:sysCanon}) is nonlinear, with $A=A(\xi, t)$ and $B=B(\xi, t)$ in general.
However, in a neighborhood of an equilibrium configuration with
$\bar\xi = (\bar {q}_h, \bar{q}_r)$ and zero velocity ($\bar u = 0$),
a linear approximation of system \eqref{eq:sys1} is simply
obtained by substituting the constant matrices $J_{he}(\bar{q})$,
$J_{re}(\bar{q})$, $J_{hc}(\bar{q})$, and $J_{rc}(\bar{q})$ in $A$ and
$B$.

\subsection{Control of Physically Connected Human--Robot Systems}
We first consider here cases where the human and robot bodies are
physically connected, such as in the first three rows of
fig.~\ref{fig:startModel}. In these cases, we assume that internal
kinematic maps $\widehat{J}_{he}$ and $\widehat{J}_{hc}$ are
reasonably accurate, and can be identified with the actual Jacobians
$J_{he}$ and $J_{hc}$ of the kinematic model presented in
\eqref{eq:sys1}. These represent the Jacobians responsible for the
reaching and compensation tasks, taking into account the contribution
of the human joints. This assumption is justified with the ability of
humans to include in their body schema devices and appendages, and to
control them accurately in space, as mentioned before. Therefore,
$\widehat{J}_{he} \equiv J_{he}$, $\widehat{J}_{hc} \equiv J_{hc}$ and
$\widehat{J}_h \equiv J_h$.
In such a linearized system, the transition matrix $A$ is a
product of a projector matrix ($J_h J_h^\dagger$, where $J_h^\dagger =
J_w^{-1} J_{h}^T W$) times a block diagonal matrix of human gains
$\Lambda$, which is reasonable to assume to be positive
definite. Hence, given the minus sign in $A$, system
(\ref{eq:sysCanon}) is at least marginally stable in any $\bar
\xi$. This result corresponds to the observation that motions of humans with a
passive prosthesis are stable, with some components of the error $\xi$
going to zero, while others (notably, compensation errors) may only
reach a non-null steady state if the prosthesis does not have enough DoFs.

It is at this point that the control of the robot or prosthesis
intervenes to improve the situation, with the goal of helping the
system reach the goal ($e_e = 0$) while canceling the compensation
($e_c = 0$) and restore a comfortable posture of the human body. To do
so, one can design an output feedback controller $u = u(y)$ (i.e.,
$\dot q_r = \dot q_r(e_c)$) in the form of a linear regulator with
standard design methods.  We hence proceed to estimate the reaching
error $e_e$ (which we assume to be known to the human, but not
available to the controller) by exploiting the measured compensation
error $e_c$, with the help of an asymptotic observer as
\begin{equation} \label{eq:sysL}
    \begin{cases}
        \hat{\dot{e}} = A\hat{e}(t) + Bu(t) + L(\hat{y} - y(t)) \\
        \hat{y}(t) = C \hat{e}(t).
    \end{cases}
 \end{equation}
By suitable choice of $L$, we can stabilize the observer dynamics, and
thus obtain an estimate of the unknown intended target of human motion.
We can then design a controller $u = - K \hat{e}$, thus
obtaining an overall output-feedback stabilization that will
eventually lead to $e_e \rightarrow 0$. More details on the choice of
$L$ and $K$ will be given shortly.

Notice that complete reachability of all states may require enough
dexterity of the robot or prosthesis, and full observability may
need rich sensorization of compensation movements. If such
properties are not warranted, however, the control designer's goal
will be limited to place reachable and observable eigenvalues to a
desirable location, while leaving others in their fixed, stable
positions.

Results in fig.~\ref{fig:ReachingError} report simulations obtained in
Matlab/Simulink for the linearized system without and with the
stabilizing controller for a prosthetic user with 7 robotic DoFs.
It can be observed that the user initially reduces the reaching error
at the expense of the compensation error
(fig.~\ref{fig:CompensationError}). Indeed, the subject must perform
compensatory motions to reach the desired end-effector pose $x_e$
without being able to move the prosthetic arm. As soon as the observer
for $e_e$ reaches convergence, the robot controller kicks in and the
prosthesis is actuated in the direction of the goal, so that also the
compensation error is reduced to zero.

The observer-based control scheme presented above for the linearized
model near an equilibrium can be applied to the general, nonlinear
model (\ref{eq:sysCanon}) by an iterative relinearization scheme in the
current working point, similar to what customarily done with Extended
Kalman Filters (EKF). In other words, knowing that the joint configurations
will vary at each time step $t$ as $q_h(t)$, $q_r(t)$, the Jacobian
matrices ($J_{he} = J_{he}(t)$, $J_{hc} = J_{hc}(t)$, $J_{re} =
J_{re}(t)$, $J_{rc} = J_{rc}(t)$) and the state matrices $A = A(t)$
and $B = B(t)$ are recomputed and used for computing the controller
and estimator gains.

Accordingly, we apply a straightforward Linear Quadratic Gaussian
(LQG) regulator design by iteratively recomputing the $K$ and $L$
matrices along the executed motions. 
Using the standard EKF procedure, the Kalman estimator gain is
computed as   
\[
    L = PC^TR_{cov}^{-1},  
\] 
with $P$ the estimated state covariance matrix and $R_{cov}$ the
covariance matrix of the observation noise.  The
estimated state covariance matrix is updated as
\[
    \dot{P} = AP + PA^T - LCP + Q_{cov},    
\]
with $Q_{cov}$ the covariance of the process noise.
For the control part, we recurrently compute the controller gain $K$
so as to minimize the cost function
\[
    \int_{t=0}^{\infty} \Big( e^{T}Qe+ u^{T}Ru+2e^{T}Su \Big)\text{dt},
\]
where $Q$ is the state-cost weight matrix, $R$ the input-cost weight
matrix, and $S$ is the mixed input-state weight matrix (normally set
to zero). Fast algorithms to solve such equations and compute the $K$
and $L$ matrices in real time are readily available as open-source
code.

Although there is no formal proof of stability for the iterative
relinearization technique above described, widespread practice shows
that the scheme works well for mildly nonlinear systems,
such as the human-robot system at hand. Simulations and experimental
results reported below confirmed this.

\subsection{Control of Physically Disconnected Human--Robot Systems}
Somehow surprisingly, the proposed method can also be adapted to the
pilot-avatar scenario depicted in the last row of
fig.~\ref{fig:startModel}, where a user acts in a remote environment
through a physically disconnected avatar.

Indeed, because of the physical robot-user disconnection,
in \eqref{eq:sys1} it holds $J_{he} = 0$.
Should also the human model in \eqref{eq:ass1} include $\hat J_{he} =
J_{he} = 0$, the operator could only try to solve the compensatory
task, and not act on the robot at all. However, this contrasts with
common experience in observing pilots in virtual reality and in
tele-operation systems \cite{Betancourt:2023}, and even in commercial
videogames. Indeed human operators often move their bodies and try
(sometimes even very hard) to move the end-effector toward a task
\emph{as if} they were the avatar to which the end-effector belongs.
In other words, the pilot immerses into the remote environment through the
perspective of the robot avatar, usually by viewing the environment
through the robot's eyes (e.g. via a head mounted display (HMD)
system). This phenomenon could be regarded as a generalized version of
the {\em rubber hand} effect, a rather well studied illusion in the
psycological literature (see e.g. \cite{Beckerle:2014}).

Based on this observation, we postulate that the human internal model
$\widehat{J}_{he}$ is not equal to the real $J_{he} = 0$ in this case.
Rather, the operator uses for $\widehat{J}_{he}$ the kinematic map of
the avatar's arm {\em as if} the arm itself was attached to the
operator's body. Given that when using avatars the egocentric frame of
reference is translated in the avatar's head, this assumption is
tantamount to having an internal model of the kinematic map from the
avatar's head to the avatar's hand, imagined as controlled by the
pilot's own body replacing the avatar's. Notice explicitly that, here
as well as in the section above, the internal model $\widehat J_{he}$
considers the artificial arm frozen in the present configuration.

\section{Numeric Analysis}
\label{sec:simulations}
\begin{figure}[t!]
    {\includegraphics[width=\columnwidth]{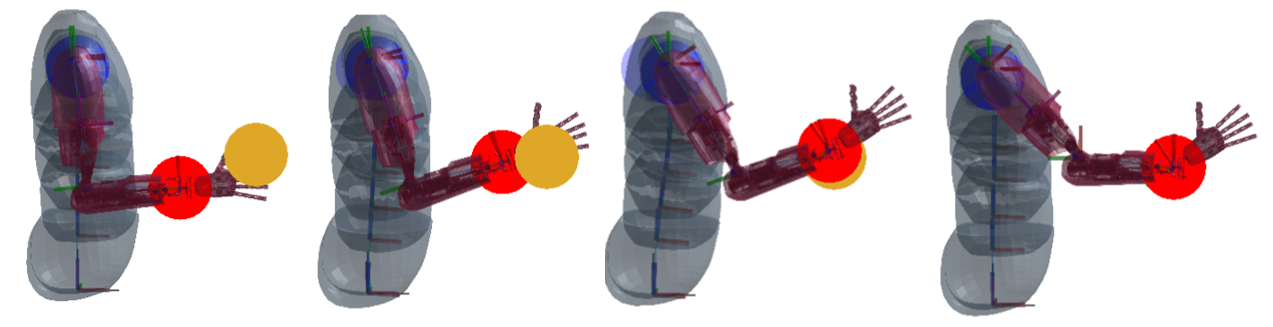}}
	\caption{Simulation 1: sequence of frames for $x_0 = \begin{bmatrix} 0, -0.5, 0 \end{bmatrix}$ $rad$ for the orientation component and  $x_0 = \begin{bmatrix} 0.2, -0.1, 0.1 \end{bmatrix}$ $m$ for the translation component. \label{fig:sim1}}
	{
		\includegraphics[width=\columnwidth]{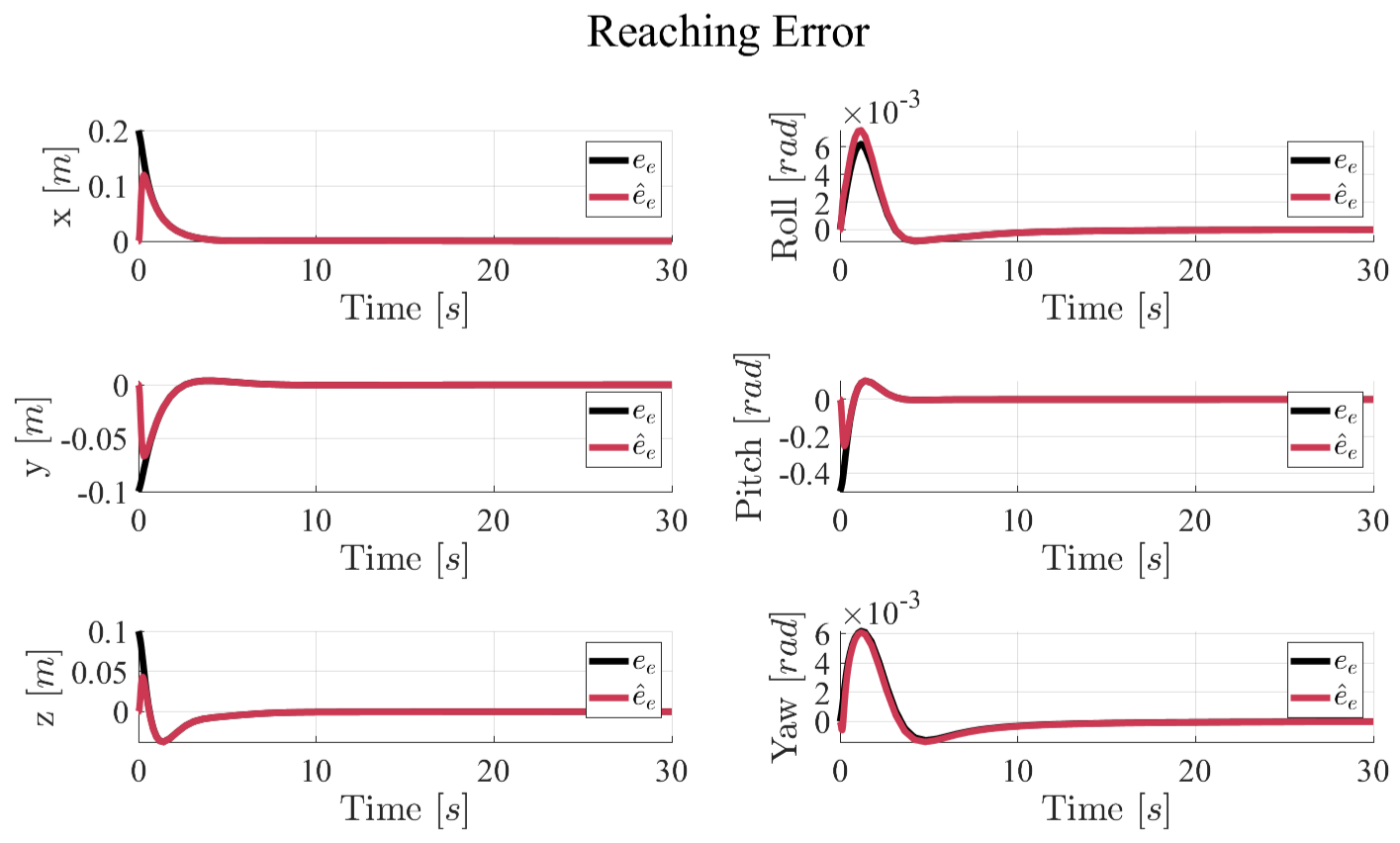}
	}
	\\[\smallskipamount]
	\centering
	\includegraphics[width = \columnwidth]{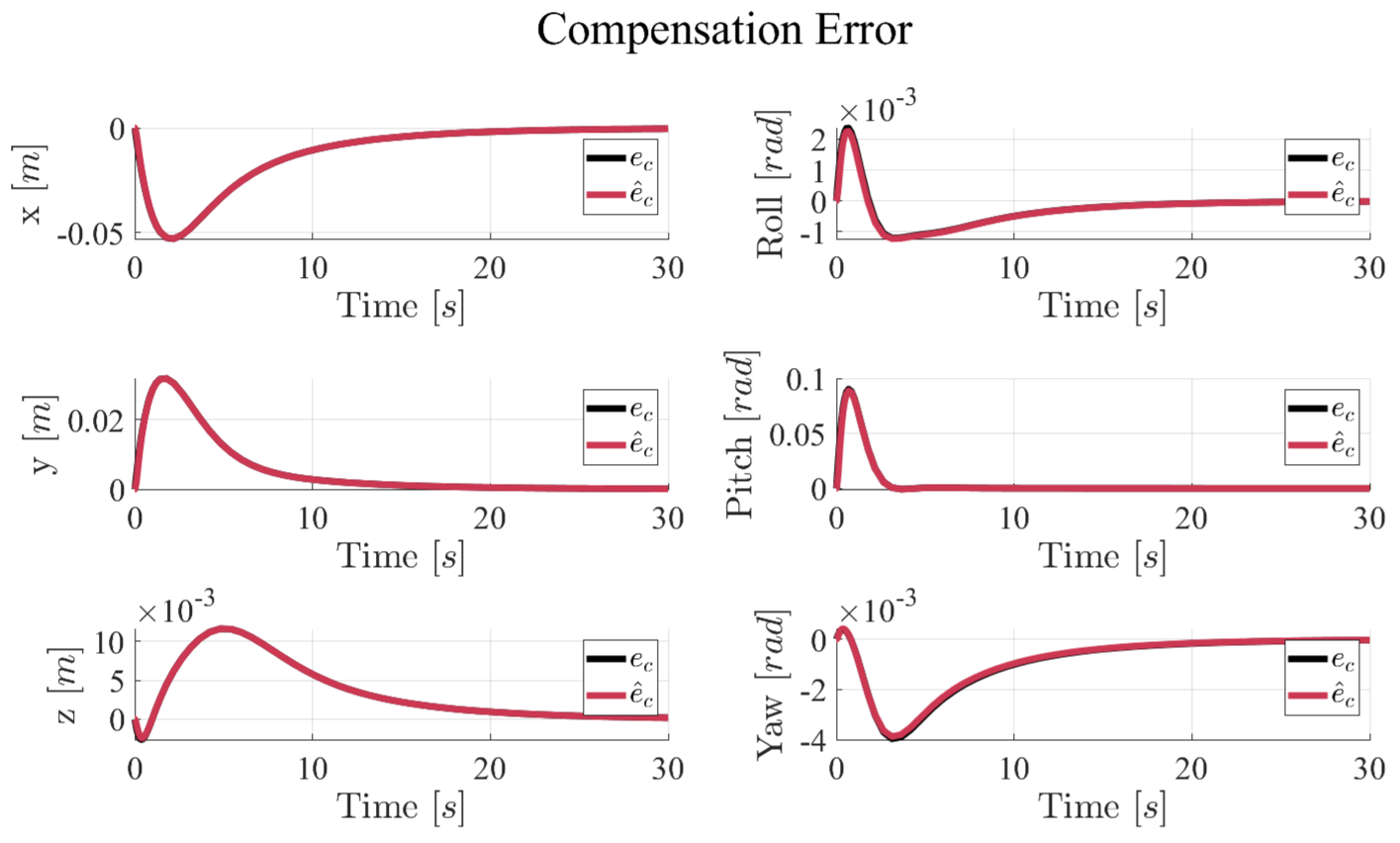}
			\caption{Simulation 1:
                            translational and rotational components of
                            reaching and compensation errors. Actual
                            values are depicted in black, their
                            EKF estimates in red. 
			\label{fig:plots_stima_sim1}}
\end{figure}

This section provides a numerical validation of the proposed control
algorithm, considering two scenarios commonly encountered in
rehabilitation robotics. The first scenario involves an upper-limb
prosthetic user, while the second features the application of the
Alter-Ego avatar robot as an assistive device. Both cases are later
replicated in the experimental section. While in the experimental
setup compensatory motion inputs will be directly taken from the human
subjects, in numerical simulations human motions are generated by our
assumed model \eqref{eq:dqh}.  Thus, this section aims to illustrate
that these assumptions lead to results consistent with expectations,
i.e. that a predefined target pose, $\bar{x}_e$, is reached, while
robot joints move so as to allow the compensatory motions to be
ultimately relaxed.  All simulations are carried out using
Matlab/Simulink.

\subsection{Simulations of Physically Connected Human-Robot System}
Our first simulations focus on the control of a transcapular
prosthesis with seven degrees of freedom. Notice that in current
prosthetic practice, no known system exists which can simultaneously
control such a complex system. In our simulated prosthesis, both the
shoulder and wrist have three DoFs, while the elbow has one.  The
compensation frame is placed at the scapula, just before the
attachment of the prosthesis.  The human behavior is modeled according
to \eqref{eq:dqh}. Following \eqref{eq:sysL}, at each time step we
calculate the observer and gain matrices $L$ and $K$ by solving the
LQG design procedure at the instantaneous system configuration.

\begin{figure}[t!]
    {\includegraphics[width=\columnwidth]{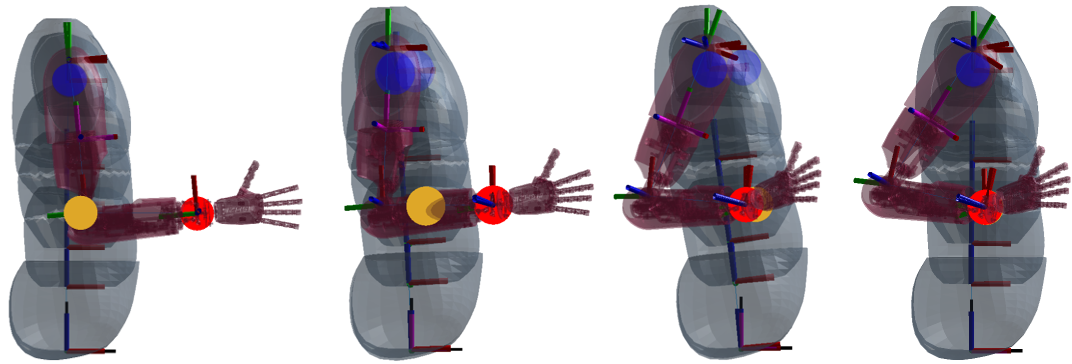}}
         \caption{ Simulation 2: sequence of frames for $x_0
          = \begin{bmatrix} -0.2, 0, -0.8 \end{bmatrix}$ $rad$ for the
          orientation component and $x_0 = \begin{bmatrix} -0.25,
            -0.1, 0 \end{bmatrix}$ $m$ for the translation
          component. \label{fig:sim2}}
	{
		\includegraphics[width=\columnwidth]{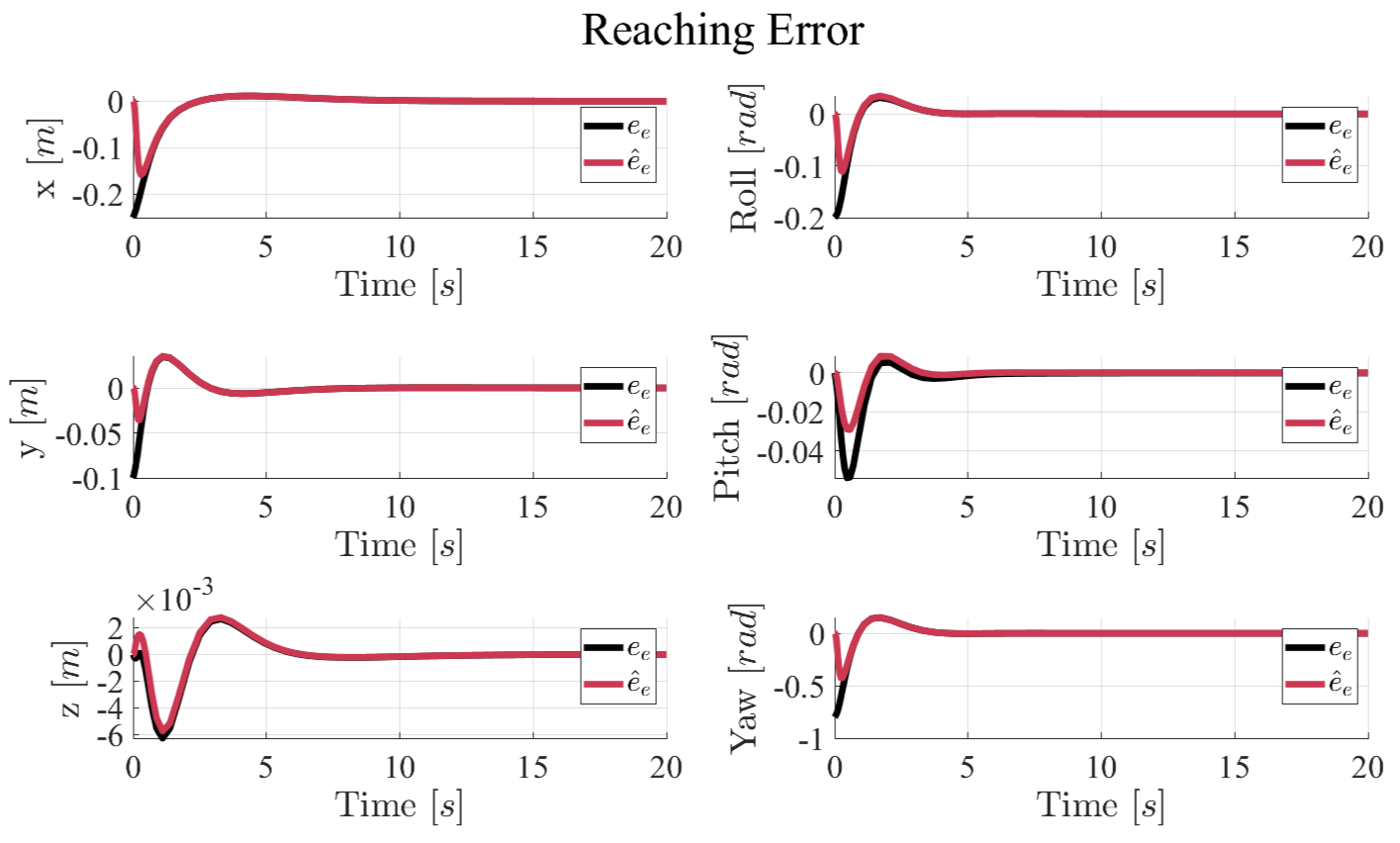}
	}
	\\[\smallskipamount]
	\centering
	\includegraphics[width = \columnwidth]{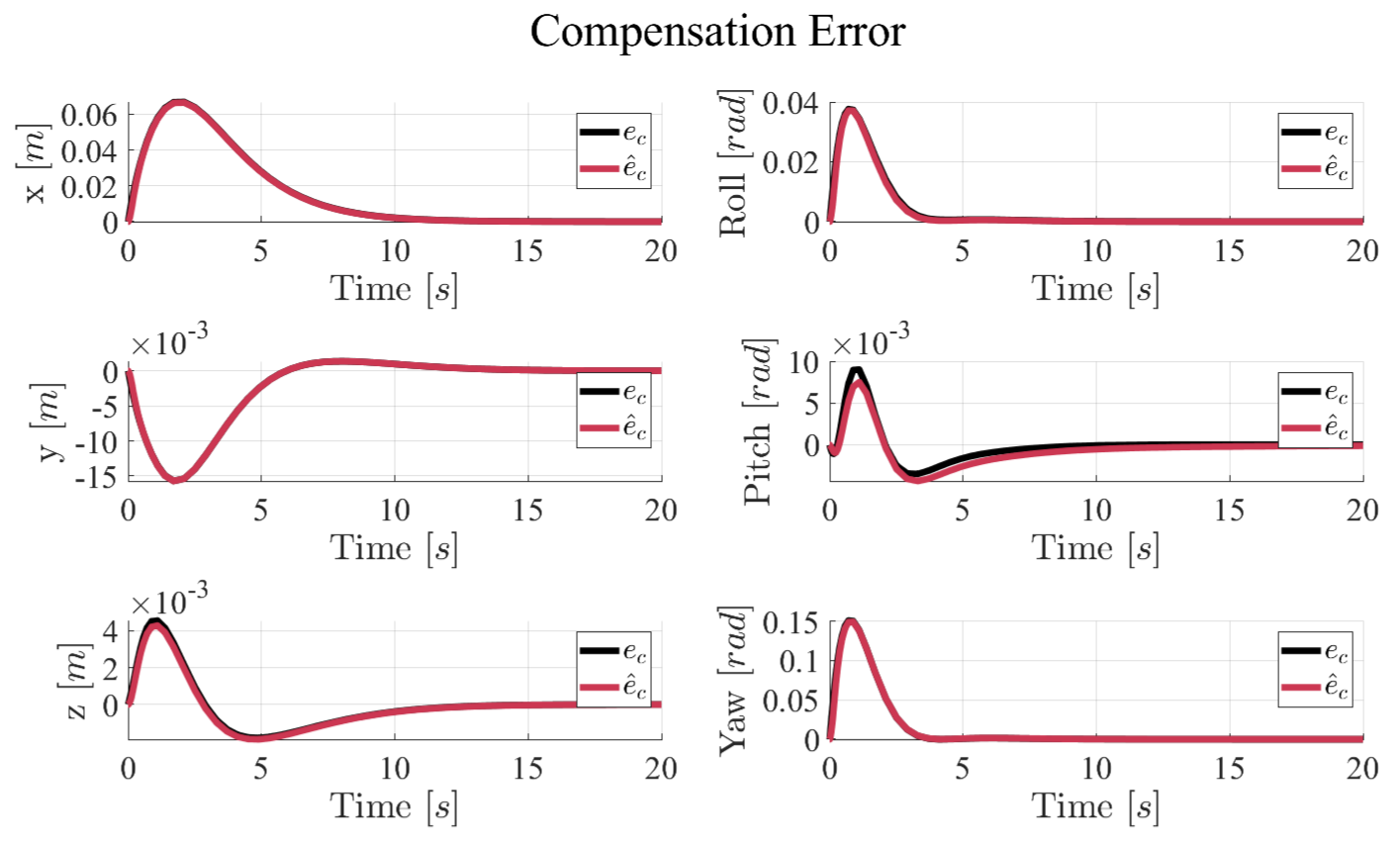}

	\caption{Simulation 2:
                            translational and rotational components of
                            reaching and compensation errors. Actual
                            values are depicted in black, their
                            EKF estimates in red. \label{fig:plots_stima_sim2}}
\end{figure}

Two simulation runs are shown for illustration, as reported in
fig.~\ref{fig:sim1} and fig.~\ref{fig:sim2}, respectively. Human body
parts are represented in gray, while artificial parts are in
purple. In both cases, the user starts from a neutral, ``home''
configuration and has to reach a desired final pose with the hand,
with both a desired position and orientation. The desired position is
farther than the home configuration in fig.~\ref{fig:sim1}, and closer
in fig.~\ref{fig:sim2}.  In both cases, our models generate first a
compensatory motion of the comensatory frame in the direction of the
target, triggering the prosthetic joints (purple). On the last step,
the prosthetic hand frame (red dot) has reached the desired target
frame (yellow dot), while the human body has returned to the relaxed
initial posture.

Fig.~\ref{fig:plots_stima_sim1} and \ref{fig:plots_stima_sim2} report
the translational and rotational components of errors, corresponding
to simulation runs of fig.~\ref{fig:sim1} and fig.~\ref{fig:sim2},
respectively. The plots report both actual (black) and estimated (red)
values of reaching and compensation errors, and show good convergence of all values to zero.

\subsection{Simulation of Physically Disconnected Human-Robot System}
The last simulation focuses on the pilot-avatar scenario, where the
user aims for the avatar to reach a predefined end-effector pose.  In
this case, the pilot assumes the perspective of the robot by viewing
the environment through the robot's eyes (e.g. by wearing a HMD
displaying the robot's camera outputs).

As described in a previous section, we consider in this scenario the
Jacobian matrix $\widehat{J}_{he}$ to represent the map from the human
reference frame (identified with the avatar's own) to the end-effector
of the artificial arm.

The derivation of the human controller and the controller follows as
above, where however the system matrix $A$ for this case is
\begin{equation} \label{eq:Jhe_ego}
  A = - J_h \bar J_w^\dagger \widehat{J}_h W \Lambda, 
\end{equation} 
with $J_h = \begin{bmatrix} 0 \\ J_{hc} \end{bmatrix}$ and $\widehat{J}_h = \begin{bmatrix} \widehat{J}_{he} \\ J_{hc} \end{bmatrix}$. 

\begin{figure}[b!]
    {\includegraphics[width=\columnwidth]{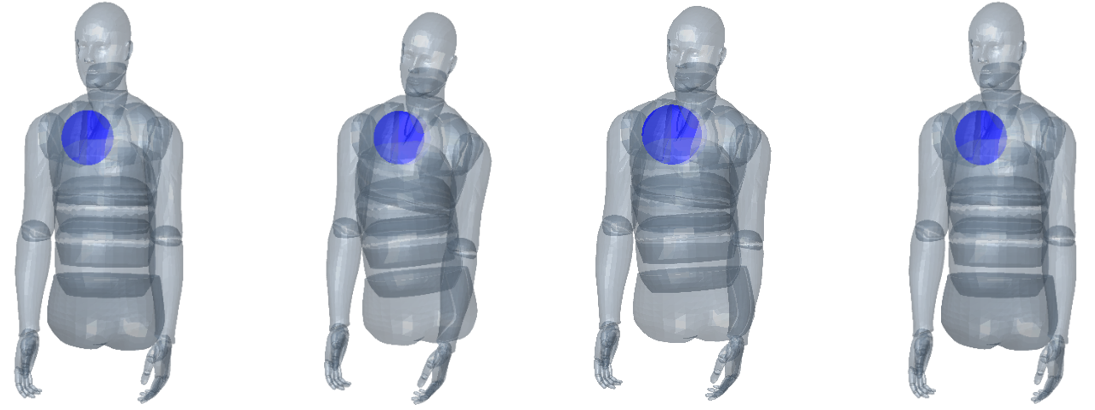}}
	\caption{Simulation 3: compensatory behavior of the user. The
          starting conditions for the reaching error are set equal to
          $x_0 = \begin{bmatrix} 0, -0.5, 0 \end{bmatrix}$ $rad$, for
          the orientation and $x_0 = \begin{bmatrix} 1, 0,
            0.1 \end{bmatrix}$ $m$ for the translation. The blue dot
          placed on user's shoulder identifies the compensation
          frame. \label{fig:sim_ego_human}}
		  \includegraphics[width = \columnwidth]{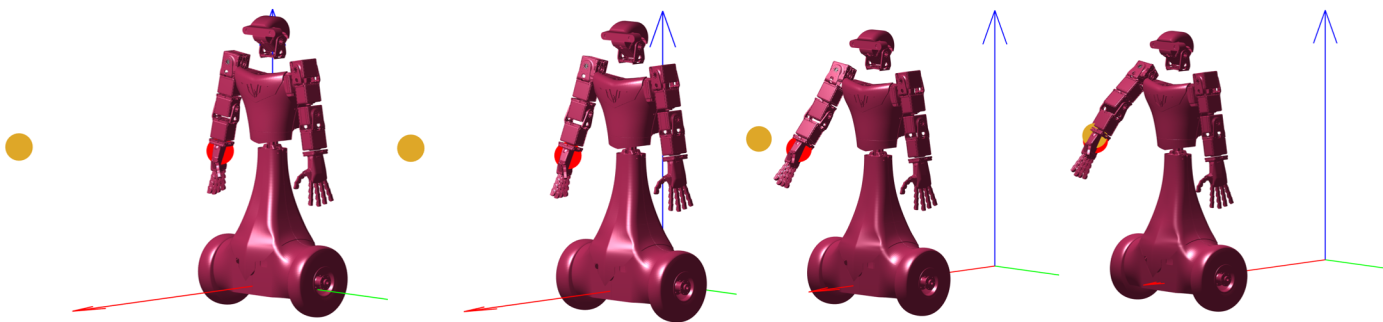}
		  \caption{Simulation 3: reaching motion of the avatar. The robot end
				effector (red dot) reaches the desired pose (yellow dot)
				through coordinated control of base and arm. \label{fig:sim_ego_avatar}}
\end{figure}

For this study, we employed the Alter-Ego robot \cite{Zambella:2019},
a two-wheeled auto-balancing humanoid avatar. The robot has a mobile
base for translation and rotation, along with two robotic arms, each
with 5 DoFs, attached to the main robotic body.  Considering its
kinematics, we designated the matrix $J_{re}$ as the Jacobian matrix
of the entire robotic chain from the global frame (fixed to the
ground) to the right end effector, thus being influenced by the joints
of the wheels and the right arm.  The robotic base is described by the
$x$ and $y$ coordinates of the center of its base, and by the orientation
angle $\theta$ around the vertical axis $z$. The base motions are
modeled as a unicycle, with the translation and rotation velocities as
inputs.

\begin{figure}[t!]
	 {
          \includegraphics[width=\columnwidth]{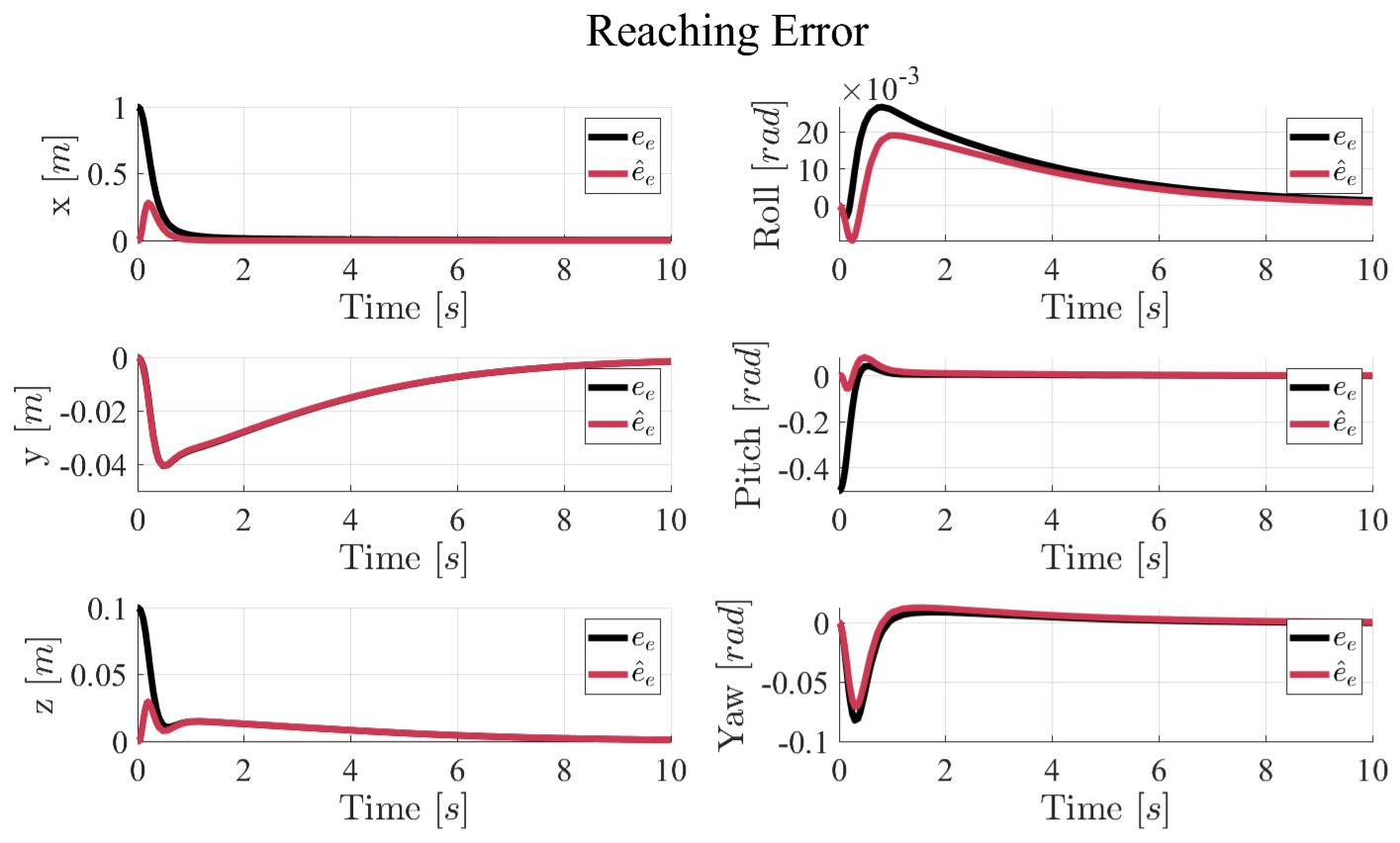}
        } \\[\smallskipamount] \centering \includegraphics[width =
          \columnwidth]{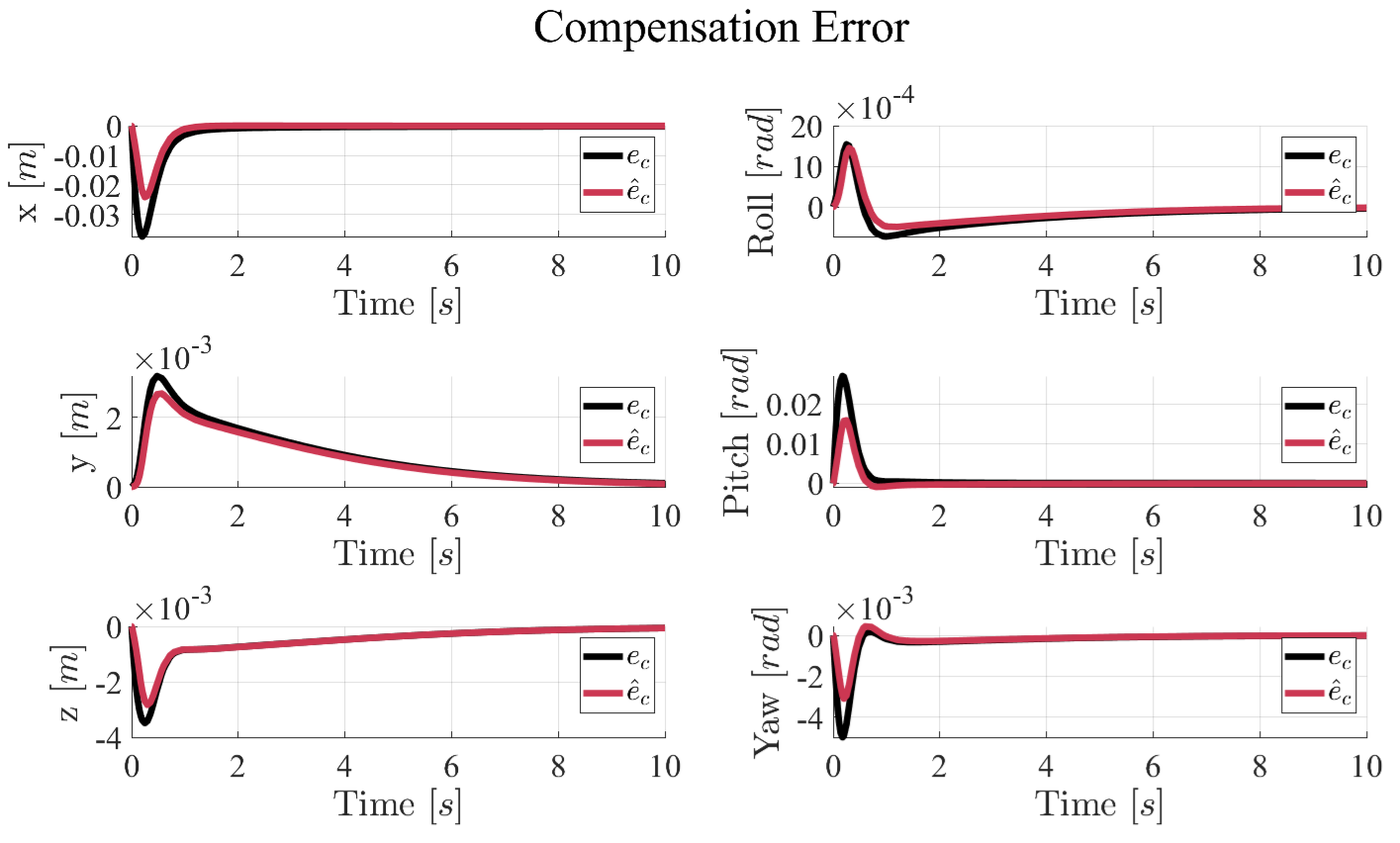}

			\caption{Simulation 3: translational and rotational components of
                            reaching and compensation errors. Actual
                            values are depicted in black, their
                            EKF estimates in red. \label{fig:plots_stima_sim3}}
\end{figure}

The simulation was implemented in Matlab/Simulink by specifying a
desired reaching pose for the robot's right end--effector. The human
motion is simulated by the model (\ref{eq:dqh}), while the EKF and
controller are computed based on \eqref{eq:Jhe_ego}.  The sequence of
frames for this simulation is depicted in
fig.~\ref{fig:sim_ego_human} and
fig.~\ref{fig:sim_ego_avatar}. Convergence of both reaching and
compensation errors, and estimates thereof, are reported in
fig.~\ref{fig:plots_stima_sim3}.

\subsection{Stability Margins}
It should be noticed that the regulator design uses estimates of the
unknown human gain matrices $\Lambda_e$ and $\Lambda_c$. These
matrices are subject-dependent, and can be estimated by means of
experiments conducted directly on each subject (see below in the
experimental section). It is to be expected
in any case that even their customized estimate will not exactly
reflect reality (including here the likely fact that the linear
relationships assumed in \eqref{eq:ass1} and \eqref{eq:ass2} may not
be valid except for a neighborhood of the equilibrium). In reality,
then, we will have to deal with a perturbed system model, whereby the
computation of the regulator matrices uses an estimate of the system
dynamics.

\begin{figure}[t!] 
    {\includegraphics[width = \columnwidth]{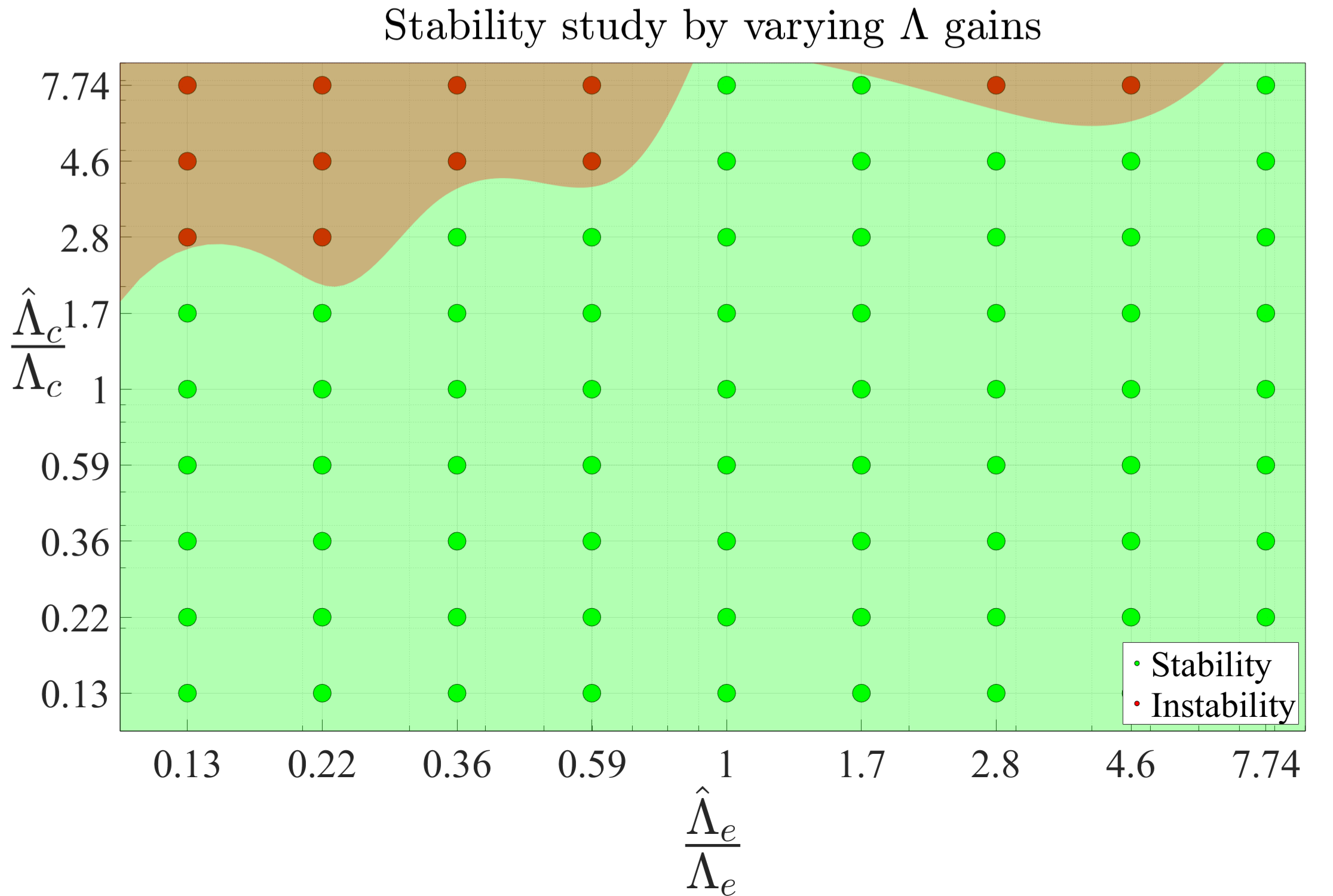}}
	\caption{Stability
   of the system as the ratios of estimate and real human model
   parameters, $\frac{\hat{\Lambda}_e}{\Lambda_e}$ and
   $\frac{\hat{\Lambda}_c}{\Lambda_c}$, are varied. Green dots
   represent relative errors for which stability is maintained, while
   in red dots instability occurs. \label{fig:lambda}}
\end{figure}

Establishing the limits for the estimation error within which the
regulator still guarantees stability of the system is a problem in
robust control which is not studied here. We note however that, even
if such bounds could be established mathematically, their application
to the case at hand would not be not trivial without a thorough
psychophysical analysis.  To provide a glimpse into the robustness
properties of the method, however, we report in fig.~\ref{fig:lambda}
results of a few simulations for the linearized system of a
prosthetic user with 7 robotic DoFs, where we assume the true human
control parameters to be normalized at $\Lambda_e = \Lambda_c = 1\, I_6$,
and their estimates $\hat{\Lambda}_e$ and $\hat {\Lambda}_c$ to vary
in the range $\begin{bmatrix} 10^{-2}\, I_6, 10^{2}\, I_6 \end{bmatrix}$. For each
simulation test, the behavior of reaching and compensation errors is
simulated. In the plot, we highlight in green the values of $\frac{\hat{\Lambda}_e}{\Lambda_e}$ and
$\frac{\hat{\Lambda}_c}{\Lambda_c}$, for which a stable behavior is
maintained.

\section{Experiments and Results}
\label{sec:experiments}
We validate our approach through two experimental scenarios designed
to assess the algorithm's performance across several robotic degrees
of freedom. We include a video in the attachment of this study showing both simulated and experimental implementations. We implement the two experiments in the Robot Operating
System (ROS) Noetic framework and use MVN XSens 2021.2 as Inertial
Measurement Units (IMUs) to track the user's {upper-body} compensatory
motions.  In both cases, the user performs compensatory motions, which
are read as human joint angles and processed by the central ROS
node. We define the user-dependent kinematics for computing the
Jacobian matrices according to the Unified Robot Description Format
(URDF) conventions. The final output consists of the prosthetic joint
positions.  The following subsections provide a detailed description
of the principal outcomes obtained in the experimental sessions.

\begin{figure}[b!]
	\centering
	\includegraphics[width = 50mm]{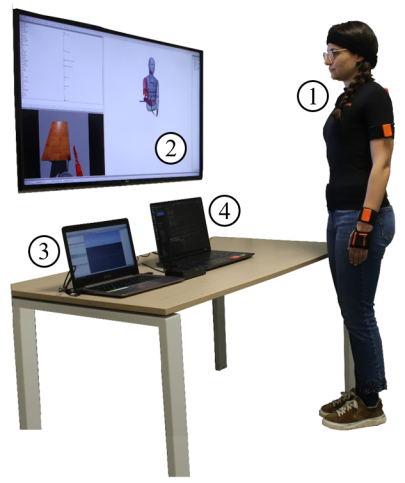}
	\caption{Experimental setup. 1: user tracked via XSens sensors. 2: Monitor for visualization of the virtual twin. 3: PC for XSens tracking. 4: PC with controller.\label{fig:vr_exp}}

	\centering
	\includegraphics[width = 7cm]{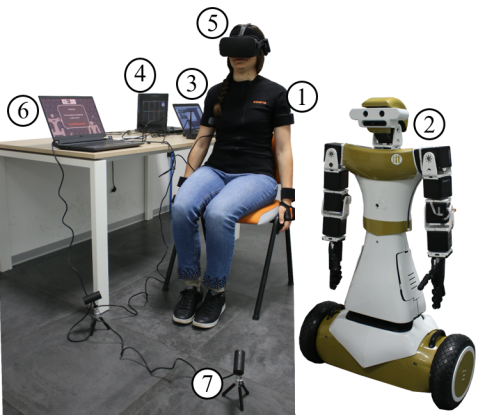}
	\caption{Experimental setup. 1: user tracked via XSens sensors. 2: Alter-Ego robot. 3: PC for XSens tracking. 4: PC with controller. 5: Oculus Rift. 6: PC for Oculus software. 7: Towers for Oculus calibration.  \label{fig:ego_exp}}
\end{figure}

\subsection{Experiments with a Virtual Upper-Limb Prosthesis} 
In the first set of experiments, we validate the proposed controller
by testing it with a real user of a virtual prosthesis. To do so, we
build a kinematically accurate digital twin of the subject, and
replicate motions captured from the subject in the twin in real
time. Virtual prostheses, with 3, 4 and 7 DoFs respectively, are used in
different experiments.  The virtual prostheses are
virtually donned to different sections of the intact subject's
arm. The amputation level is known to the experimenter and is used to
set the human and prosthesis models in the controller.

The user wears IMU sensors on their upper body and looks at their
virtual twin on a screen. A picture of the setup used for this
experiment is reported in fig. \ref{fig:vr_exp}. The subject is
instructed to reach the desired object, depicted as a red sphere in
the virtual environment by moving the attachment section of the
virtual prosthesis in their own body. The subject is instructed not to
move the parts of their real arm corresponding to the prosthesis in
use. Even if instructions are not followed, however, motions of these
parts are filtered and discarded as they are irrelevant to the
controller.

Compensatory motions of the user are tracked and mapped onto the
virtual twin. The algorithm interprets compensatory motions to
directly control the robotic joints of the virtual prosthesis. 

\begin{figure}[t!]
	\centering
	\includegraphics[width = \columnwidth]{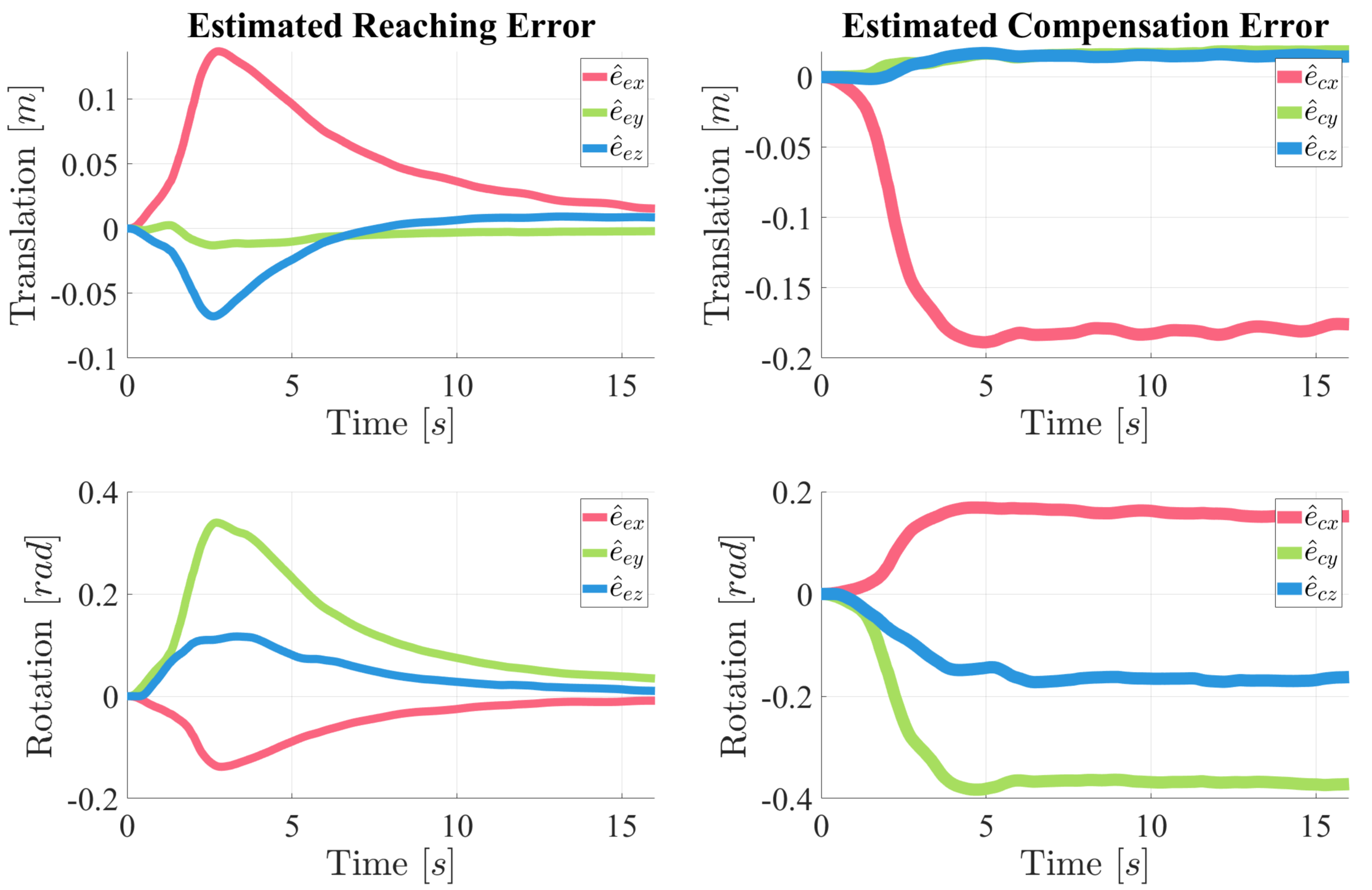}
	\caption{{7 DoFs user with the prosthetic controller turned off: estimation of the reaching error $\hat{e}_e$ (left) and of the compensation error $\hat{e}_c$ for the case of the prosthetic controller turned off.} \label{fig:PassiveProst}}

	\centering
	\includegraphics[width = \columnwidth]{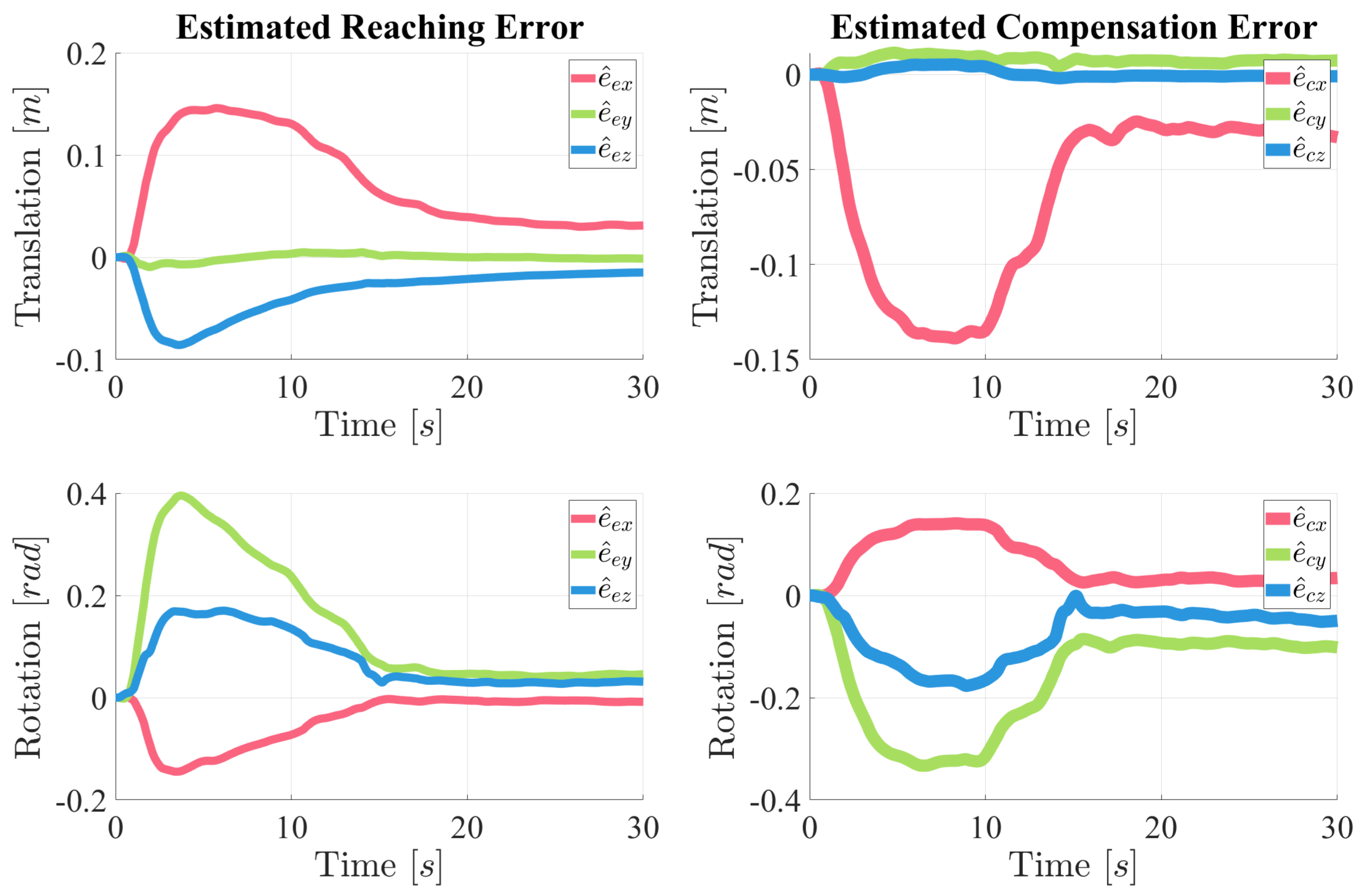}
	\caption{{7 DoFs user with the prosthetic controller turned on: estimation of the reaching error $\hat{e}_e$ (left) and of the compensation error $\hat{e}_c$ for the case of the prosthetic controller turned on.} \label{fig:PassiveProstControl}}
\end{figure}

\begin{figure*}[t!]
    \centering
    {\includegraphics[width=\textwidth]{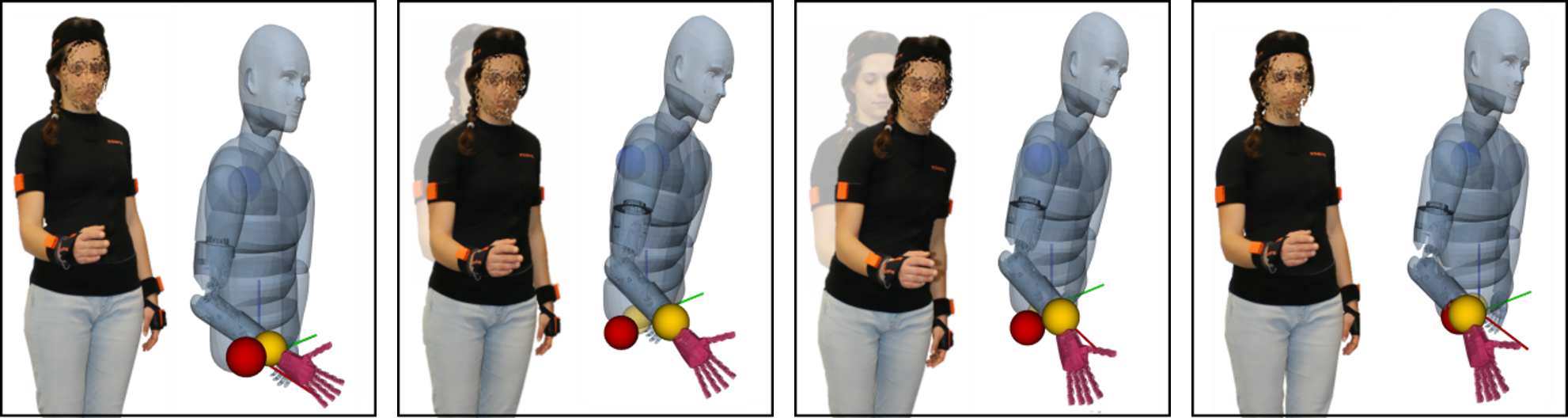}}
	\caption{Scenario 1: sequence of frames for the transradial amputation case, with 3 prosthetic DoFs. \label{fig:seqVir_3DoF}}
\end{figure*}

\begin{figure*}[t!]
    \centering
    {\includegraphics[width=\textwidth]{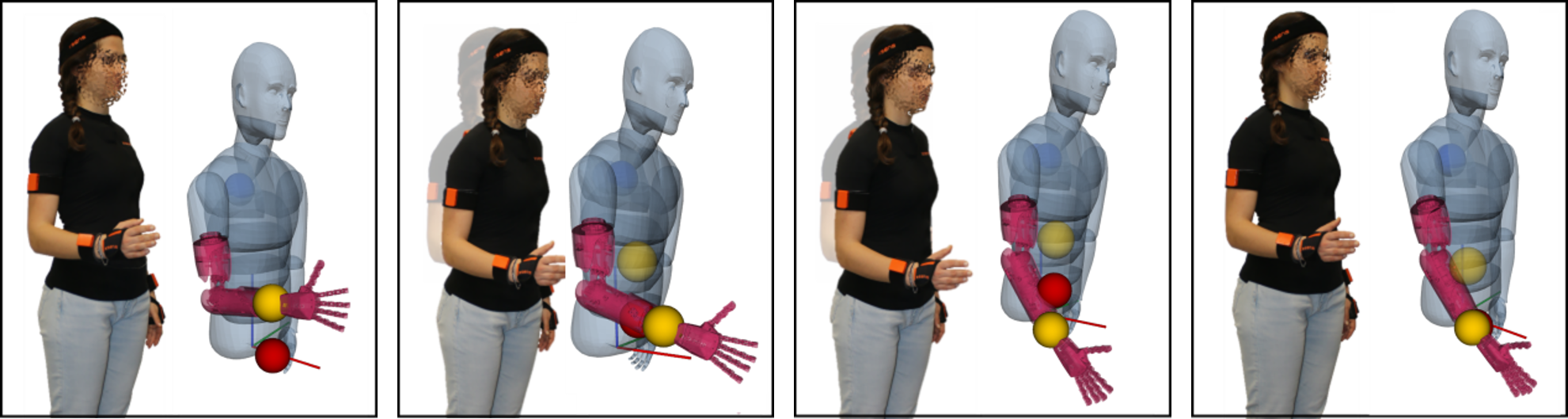}}
	\caption{Scenario 2: sequence of frames for the transhumeral amputation case, with 4 prosthetic DoFs. \label{fig:seqVir_4DoF}}
\end{figure*}

\begin{figure*}[t!]
    \centering
    {\includegraphics[width=\textwidth]{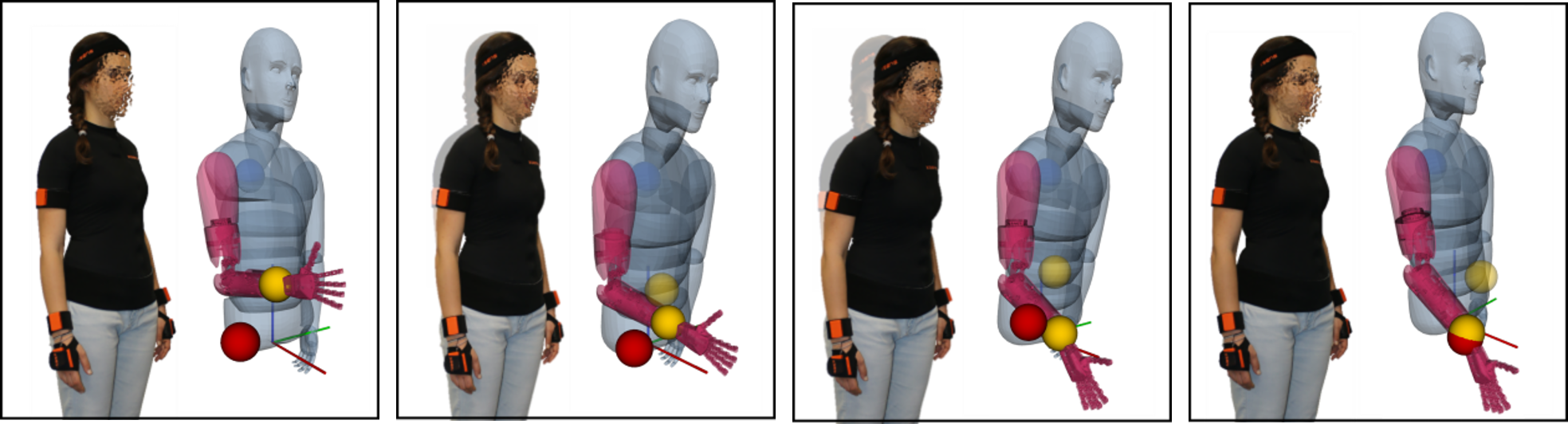}}
	\caption{Scenario 3: sequence of frames for the shoulder disarticulation case, with 7 prosthetic DoFs. \label{fig:seqVir_7DoF}}
\end{figure*}

\begin{figure*}[t!]
    \centering
    {\includegraphics[width=\textwidth]{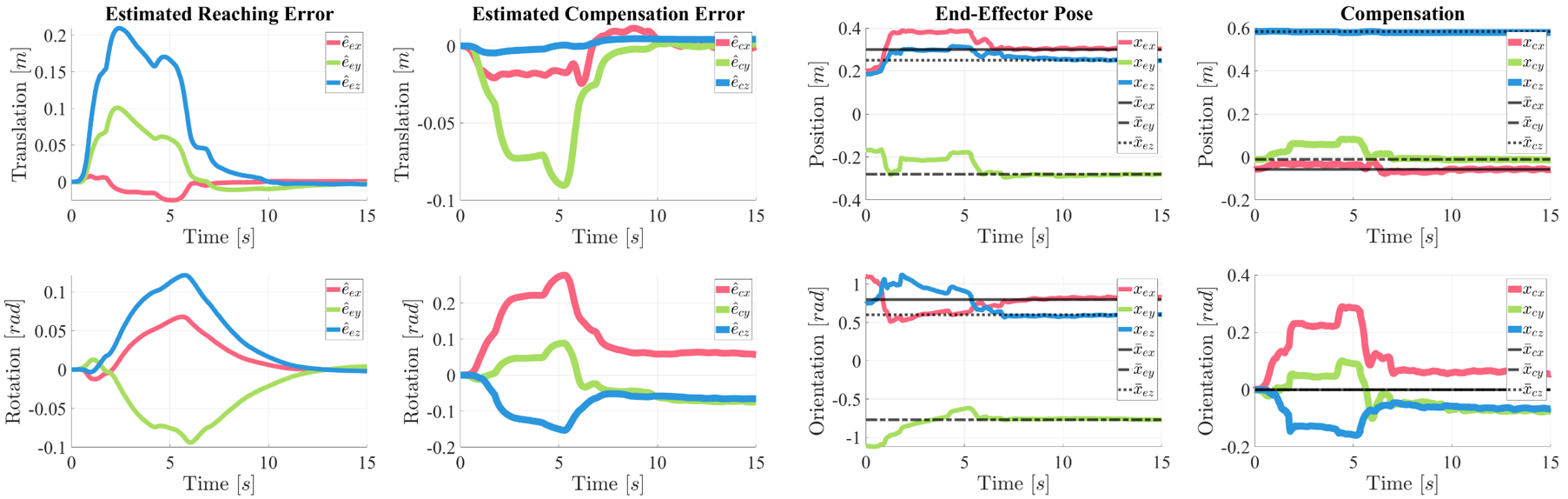}}
	\caption{{Scenario 1: estimation of the reaching and compensation errors and actual reaching and compensation frames for the transradial amputation case.} \label{fig:3dof_plots}}
\end{figure*}

\begin{figure*}[t!]
    \centering
    {\includegraphics[width=\textwidth]{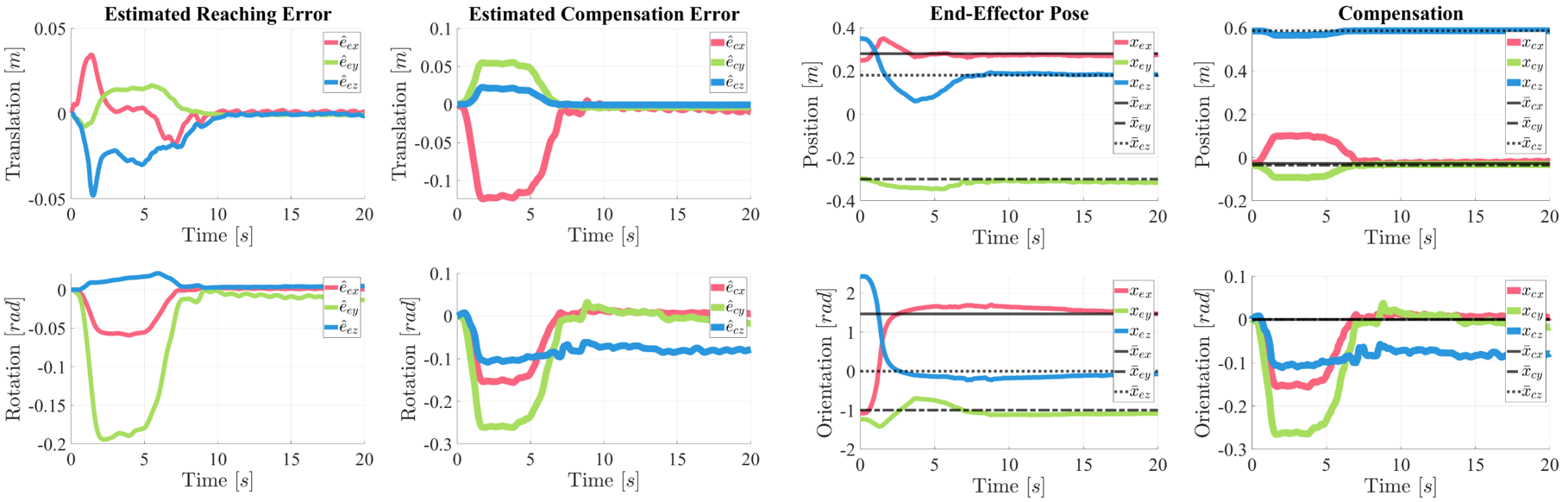}}
	\caption{Scenario 2: estimation of the reaching and compensation errors and actual reaching and compensation frames for the transhumeral amputation case. \label{fig:4dof_plots}}
\end{figure*}

\begin{figure*}[t!]
    \centering
    {\includegraphics[width=\textwidth]{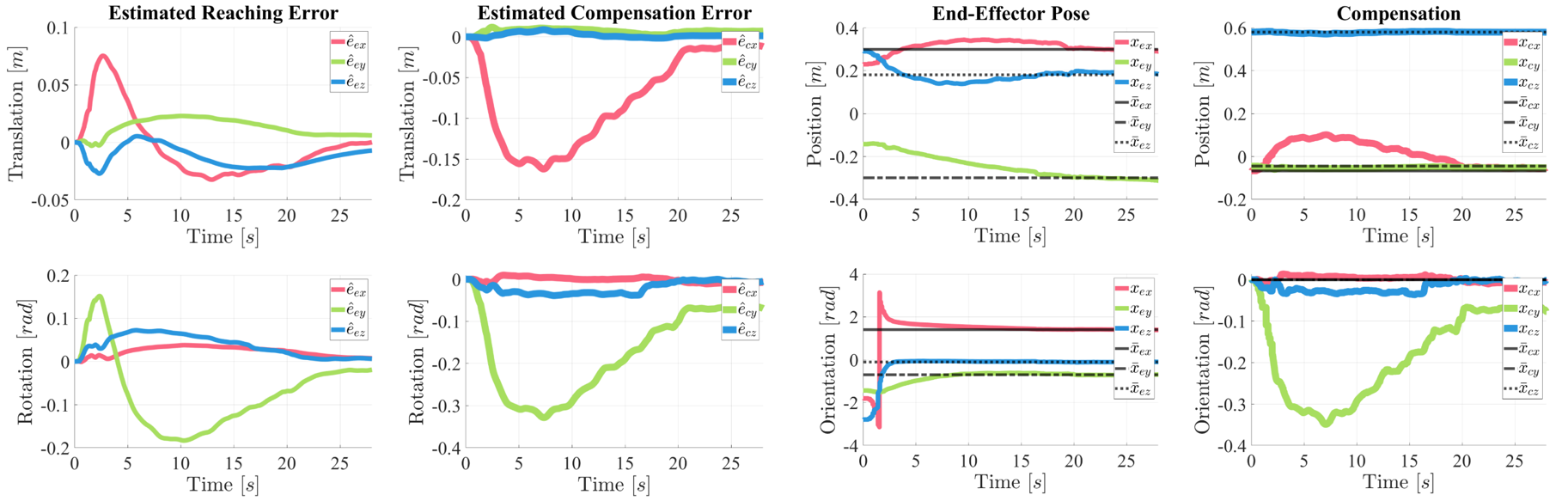}}
	\caption{{Scenario 3: estimation of the reaching and compensation errors and actual reaching and compensation frames for the shoulder disarticulation case.} \label{fig:7dof_plots}}
\end{figure*}

{\bf Remark.}  The choice of where the compensatory frame should be
placed requires some consideration. In cases where the artificial limb
is a prosthesis of a distal part of the upper limb, for instance, one
could ask whether the most distal section of the body available for
IMU instrumentation should be used as compensatory frame.  However, if
the compensatory frame is placed distally, e.g. on the forearm stump
for a transradial amputation, then the feedback policy of bringing
$e_c \rightarrow 0$ would imply that the user eventually brings their
forearm to ``home'' (rest) configuration. Thus, the functional parts
of the arm would not participate in the reaching task, which would
have to be solved by the prosthesis alone. By placing the compensation
frame at the user's shoulder instead, the final configuration will see
the partly human, partly artificial upper limb joints both participate
in reaching, while the shoulder and torso will eventually return to
rest pose. We use this solution in our experiments below.  However, it
should also be emphasized that in other scenarios than the prosthetic
arm case, and with different users, the choice could be
different - as e.g. with robot avatars for patients with very limited
residual mobility and/or strength, where instead compensation frames
could be placed at the most distal available body parts.

In a first set of experiments, we disabled the robot prosthesis
controller (i.e., we set $\dot{q}_r = 0$), and only focused on the
model of the human compensatory behaviour.
Through a series of repeated experiments and an empirical optimization,
the parameters of the subject's model were estimated to $\Lambda_c =
0.1\, I_6$, $\Lambda_e = 2\, I_6$.    

With the controller turned off and a 7 DoF arm fixed in the initial
configuration, large compensatory motions are expected to be needed to
reach the target. Results depicted in fig.~\ref{fig:PassiveProst}
highlight how the reaching error estimate (depicted on the left side)
converge to zero. As expected, the reaching target is obtained at the
expenses of the compensation error (on the right side), which remains
large. Notice that, although we do measure compensation errors, we
report here their estimates, which practically amounts to a filtered
versions of the IMU sensor outputs via the Kalman filter used in our
scheme.

In a subsequent experiment, the same target is approached with the
robot controller active, so that the prosthetic joints assist the user
in reaching the desired end-effector pose and reduce the compensatory
shoulder displacement, as depicted in
fig.~\ref{fig:PassiveProstControl}. It can be observed that
compensation movements are similar at the beginnning with the previous
case with the controller turned off, but are quickly reduced by the
prosthesis controller intervention. Convergence of the reaching error
estimates is shown to have similar dynamics as in the previous case.
A number of experiments of this type was used to tune the controller
parameters as and $P(0) = \mbox{ diag }[10, 10, 10, 10, 10, 10, 0.05, 0.05,
0.05, 0.05, 0.05]$, $R_{cov} = 0.01 \, I_6$, $Q_{cov} = I_{12}$,
$R=I_7$, and $S=0$. Finally, $Q
= \mbox{ diag }[0,0,0,0,0,0,10,10,10,0.1,0.1,0.1]$ was chosen, reflecting the
different scaling of translational and rotation components (measured in
meters and radians, respectively).

With values obtained by calibration and tuning, we performed more
experiments for three different prostheses corresponding to different
amputation levels.  The first case involves a transradial amputation
requiring control over 3 DoFs (fig.~\ref{fig:seqVir_3DoF}), a
transhumeral case follows with 4 DoFs (fig.~\ref{fig:seqVir_4DoF}),
and the final scenario refers to a shoulder disarticulation with 7
DoFs (fig.~\ref{fig:seqVir_7DoF}). In
figs. \ref{fig:seqVir_3DoF}, \ref{fig:seqVir_4DoF}, and
\ref{fig:seqVir_7DoF}, user's compensatory motions are visualized by
showing the reference configuration in transparency, and the
prosthetic motions in the virtual environment. The blue dot represents
the compensatory frame, the yellow dot is at the actual reaching frame
$x_e$, and the red dot is at the desired frame $\bar{x}_e$. Blue and
yellow dots in transparency depict the compensatory and reaching frame
respectively at the starting configuration.
As it can be seen, as soon as the user starts to produce compensatory
motions (which are directly mapped into the virtual twin's shoulder
motions), the controller starts generating commands for the
prosthesis. Once the subject is satisfied with the target pose
achieved by the robotic hand, they go back to their starting posture,
stopping the prosthesis motion $\dot{q}_r = 0$.

Figs.~\ref{fig:3dof_plots}, \ref{fig:4dof_plots},
and \ref{fig:7dof_plots} present plots corresponding to the 3DoF,
4DoF, and 7DoF prosthetic user scenarios, respectively. For each case,
plots on the two left-most columns display convergence of the
estimated reaching error and compensation error, respectively, to
zero. Plots on the right side show instead the actual pose reached by
the prosthesis hand, in comparison with the desired final pose, and
the compensation frame pose, compared to the target relaxed pose.  As
shown, in all cases the robotic hand successfully reaches the desired
target, even though the controller is unaware of its pose. Once the
target is reached, the compensation is progressively relaxed.

\begin{figure}[t!]
	\centering
	\includegraphics[width = \columnwidth]{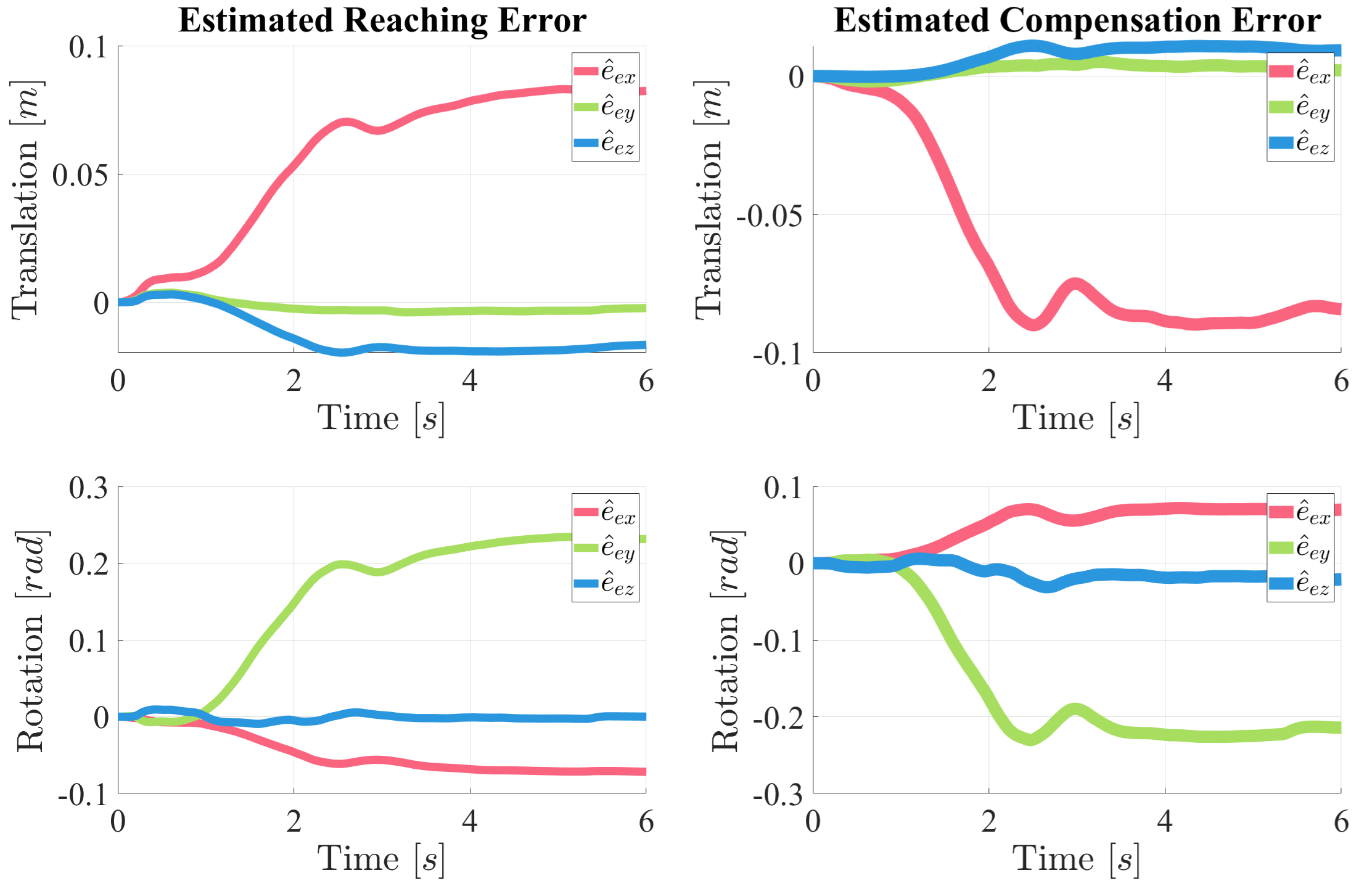}
	\caption{Pilot-avatar case with the robotic controller turned off: estimation of the reaching error $\hat{e}_e$ (left) and of the compensation error $\hat{e}_c$, for the case of the robotic controller turned off. \label{fig:PassiveEgo}}
\end{figure}

\begin{figure}[t!]
	\centering
	\includegraphics[width = \columnwidth]{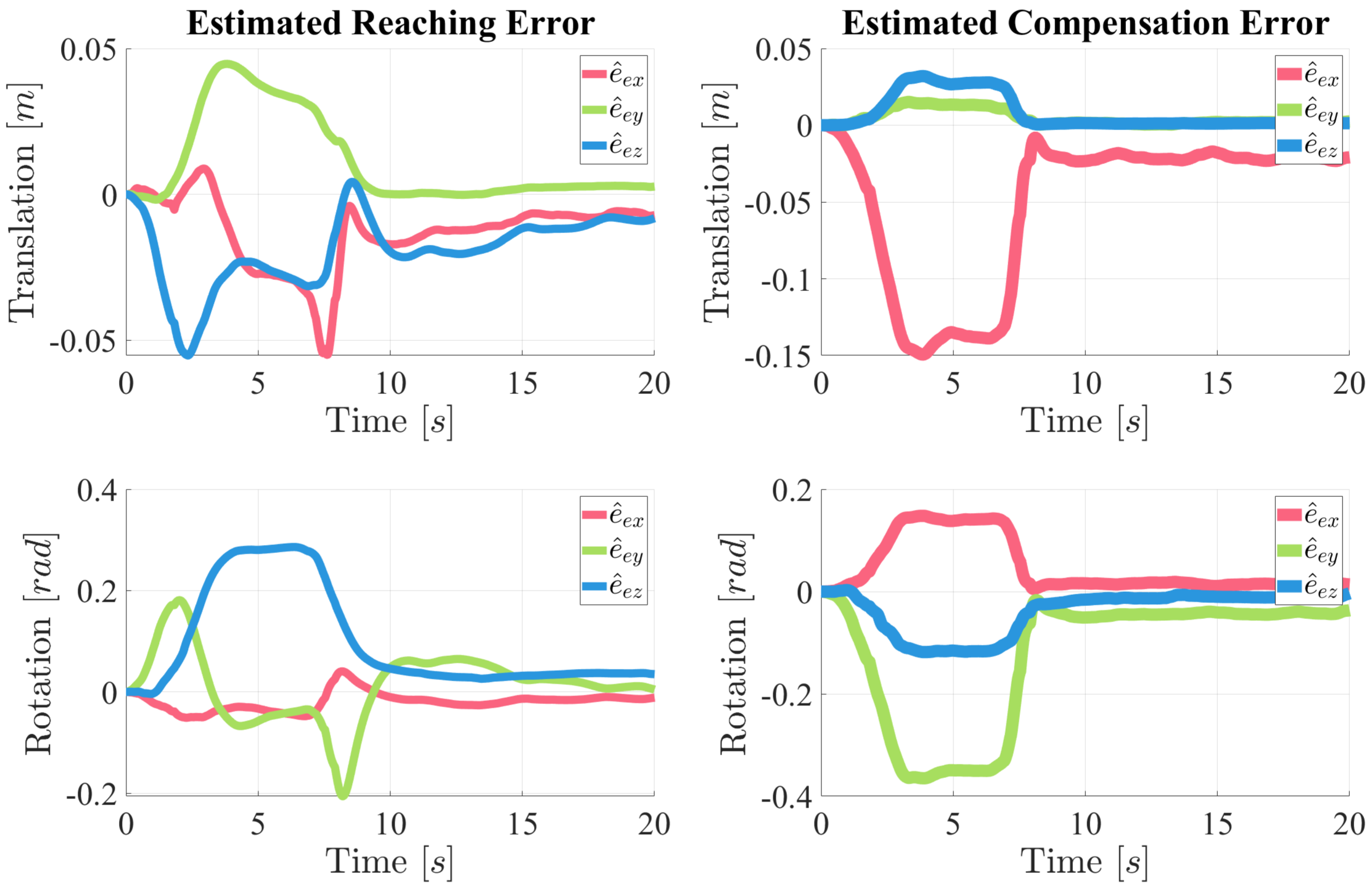}
	\caption{Pilot-avatar case with the robotic controller turned on: estimation of the reaching error $\hat{e}_e$ (left) and of the compensation error $\hat{e}_c$, for the case of the robotic controller turned on. \label{fig:ActiveEgo}}
\end{figure}

\subsection{Experiment with a Whole-Body Prosthesis}
In the second experiment, we validate the algorithm for the
two-wheeled humanoid robotic avatar
Alter-Ego \cite{Zambella:2019}. Robotic joints to be controlled here
include not only those belonging to the arm (5 joints for the right
arm) but also the wheeled base. The base is modelled by the kinematics
of a unicycle, as described in Section \ref{sec:simulations}.
\begin{figure*}[t!]
    \centering
    {\includegraphics[width=\textwidth]{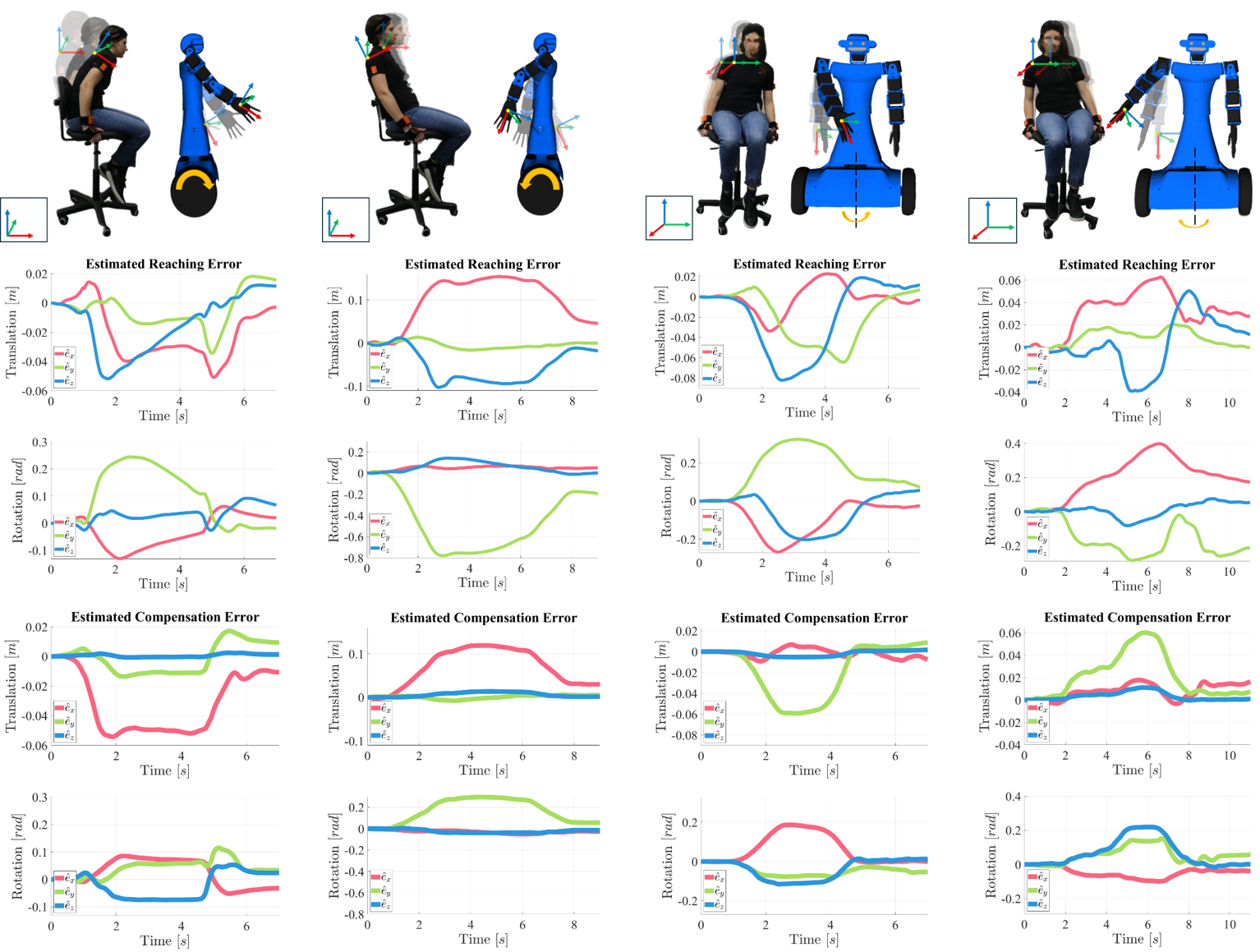}}
        \caption{Individual user motions to drive Alter-Ego
		along positive and negative directions of $x$ axis and
		positive and negative directions of $y$ axis of the
		global frame, displayed in the box on the bottom left
		corner. Both user's and robot's old frames are
		depicted in transparency. Below each case, the
		corresponding plots for both reaching and compensation
		errors are reported. \label{fig:single_motions}}
\end{figure*}
In this setup, the robot functions as a sort of ``whole-body"
prosthesis of the person, who perceives the robot avatar as an
artificial extension of their own body. The user views the environment
through Alter-Ego's eyes using an Oculus Rift as shown in
 fig.~\ref{fig:ego_exp}. This setup allows the user and the robot to be
in different physical locations.  The user's compensatory frame
motions, tracked via IMU sensors, trigger the controller. We placed
the compensatory frame at the user's right shoulder. The key
difference from the previous experiment is that, in this case, the
reaching task is entirely handled by the robot using both wheels,
shoulder and arms to reach the user's intended hand posture, without
that being known.

\begin{figure*}[t!]
    \centering
        {\includegraphics[width=\textwidth]{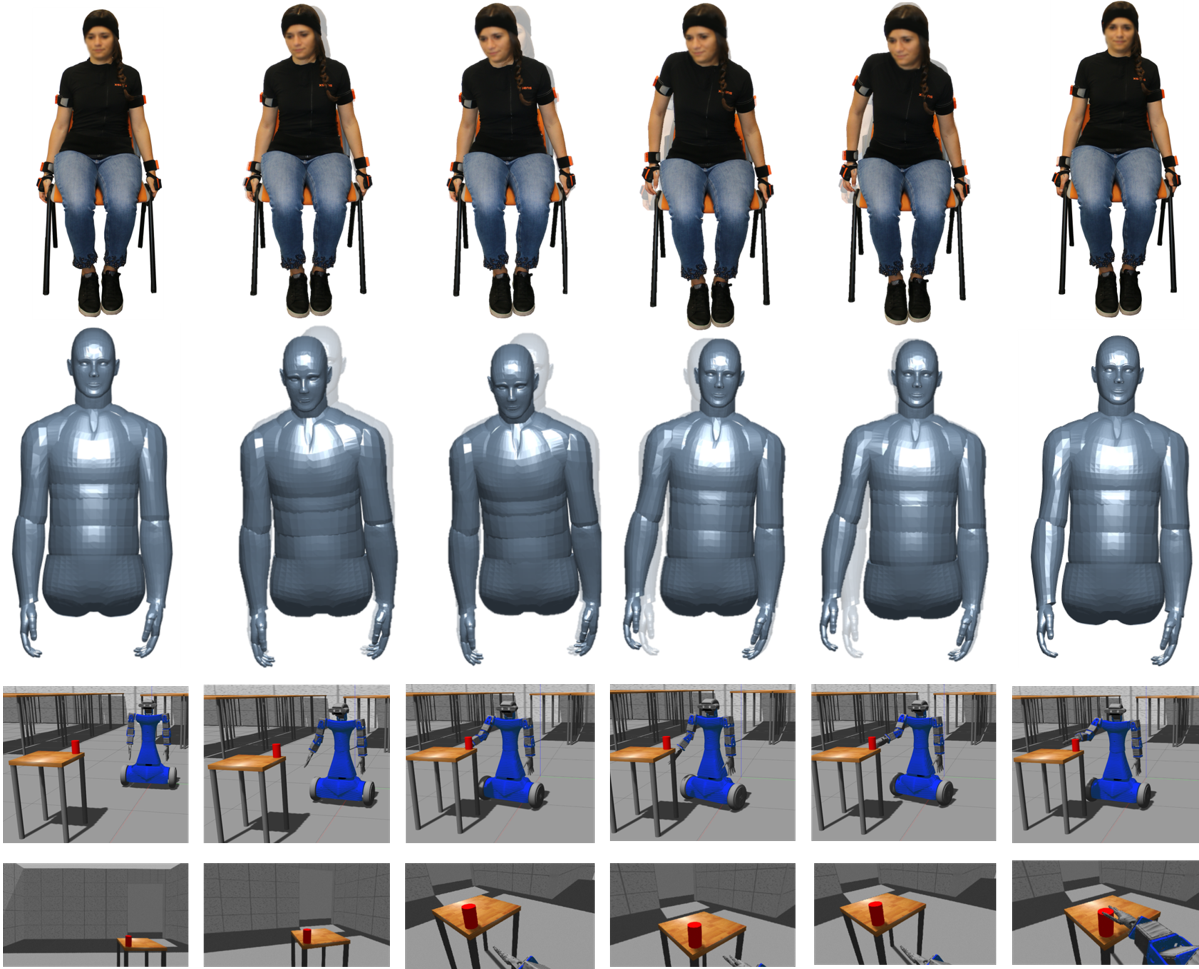}}
        \caption{Six frames of the reaching task with
		virtual model of Alter-Ego. We report the user's
		motions, tracked via XSens, the corresponding user's
		virtual twin, Alter-Ego's motion in virtual
		environment, and the view from the camera placed on
		Alter-Ego's eyes. \label{fig:seq_user_vr_reach}}
\end{figure*}

We place the reaching frame on Alter-Ego's right palm, and - similarly
as in the previous experiment described in  fig.~\ref{fig:PassiveProst}
- first verify the correct functioning of the observer while keeping
the robot still, i.e. $\dot{q}_r = 0$. As shown in
 fig.~\ref{fig:PassiveEgo}, the compensation error does not approach
zero. Different from  fig.~\ref{fig:PassiveProst}, neither the
estimated reaching error converges to zero in this case, because the
compensation does not produce any motion of the hand, being physically
disconnected.

When the controller is turned on, both reaching and compensation
errors converge to zero, as depicted in  fig.~\ref{fig:ActiveEgo}.  As
in the previous case, these preliminary experiments led to an
empirical optimization of the controller parameters, setting
$P(0) =  \mbox{ diag }[10, 10, 10, 10, 10, 10, 0.05, 0.05, 0.05, 0.05, 0.05,
0.05]$, $R_{cov} = 0.01\, I_6$, $Q_{cov} = I_{12}$, $R = I_7$, and $S=0$. 
and $Q = \mbox{ diag }[0,0,0,0,0,0,100,100,100,0.1,0.1,0.1]$. 

In a first set of experiments, we have a human user virtually
connected to a digital twin of the Alter-Ego robot in a virtual
environment.
The user's compensatory motions are tracked by XSens sensors.

Before executing a complete reaching task, we assess the functionality
of the robot control for individual user motions to observe how they
translate into the robot's actions.  As depicted in
 fig.~\ref{fig:single_motions}, we illustrate the compensatory motions
of the user along the positive and negative directions of the $x$ and
$y$ axes of the global frame (box on the bottom left corner). The $x$
axis is represented in red, and the $y$ axis in green.  The
compensation frame is consistently placed on the user's right
shoulder, and the end-effector frame on the robot's right palm.  The
user starts from an upright posture (in transparency) and starts
movements towards the direction of the object to be grasped.  For each
user motion, we observe the corresponding motion of Alter-Ego's right
arm from the home configuration (in transparency) towards the final
configuration. The yellow arrow highlights the direction of the
robot's base motion.  Under each pilot-avatar motion, we present the
corresponding estimation of the reaching and compensation errors, for
both translation and rotation components.

After using the first experiment to identify the human model and
choose the controller parameters, the system functionality is tested
on the more complex task of reaching for a red peg placed on a table
far from the robot initial position.  The peg is placed far enough
that the use of the arm only is not sufficient, and the wheeled base
must also translate and rotate to reach for the peg.  The user looks
at the virtual scene through a camera placed on Alter-Ego's eyes,
whose image is reproduced in a real display mounted on the head
of the human user. User motions are tracked by the XSens suite
sensors, and used to drive the robot controller as decribed above.

Fig.~\ref{fig:seq_user_vr_reach} shows six consecutive frames of the
execution of this task in a virtual reality environment. User's
compensatory motions, shown in the first row of
fig.~\ref{fig:seq_user_vr_reach}, are also reproduced in a virtual
twin (second row). The twin is reported here for visualization
purposes only, while it is not used in the control, nor is it seen by
the user.  The third row whose the virtual robot avatar reaching for the peg,
while the bottom row provides the views from the camera positioned on
the robot’s head and reproduced in the head-mounted display.

\begin{figure}[t!]
	\centering
	\includegraphics[width = \columnwidth]{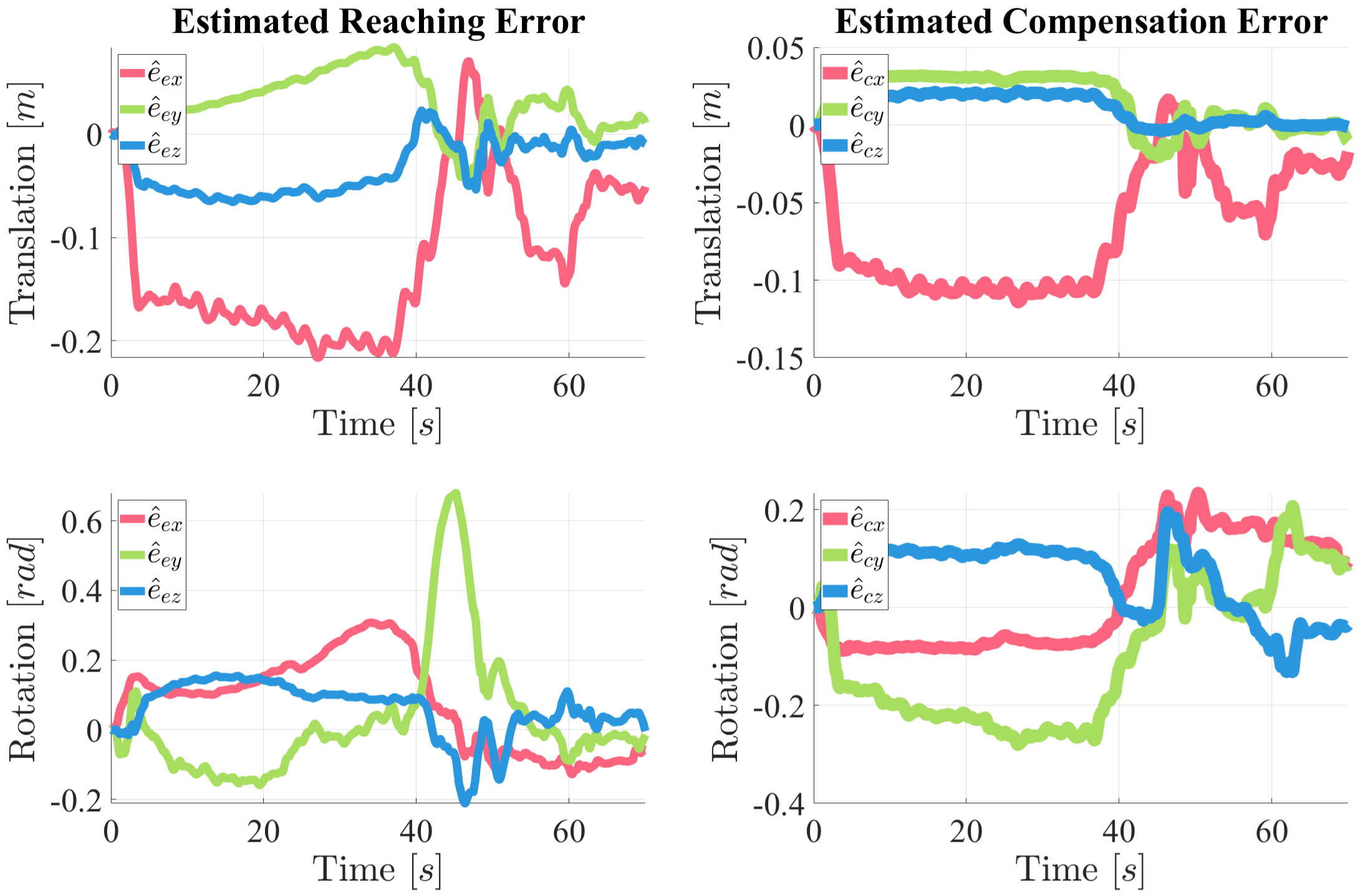}
	\caption{Estimated reaching and compensation errors for the
	VR reaching task. \label{fig:Errors_vr_ego}}

	\centering
	\includegraphics[width = \columnwidth]{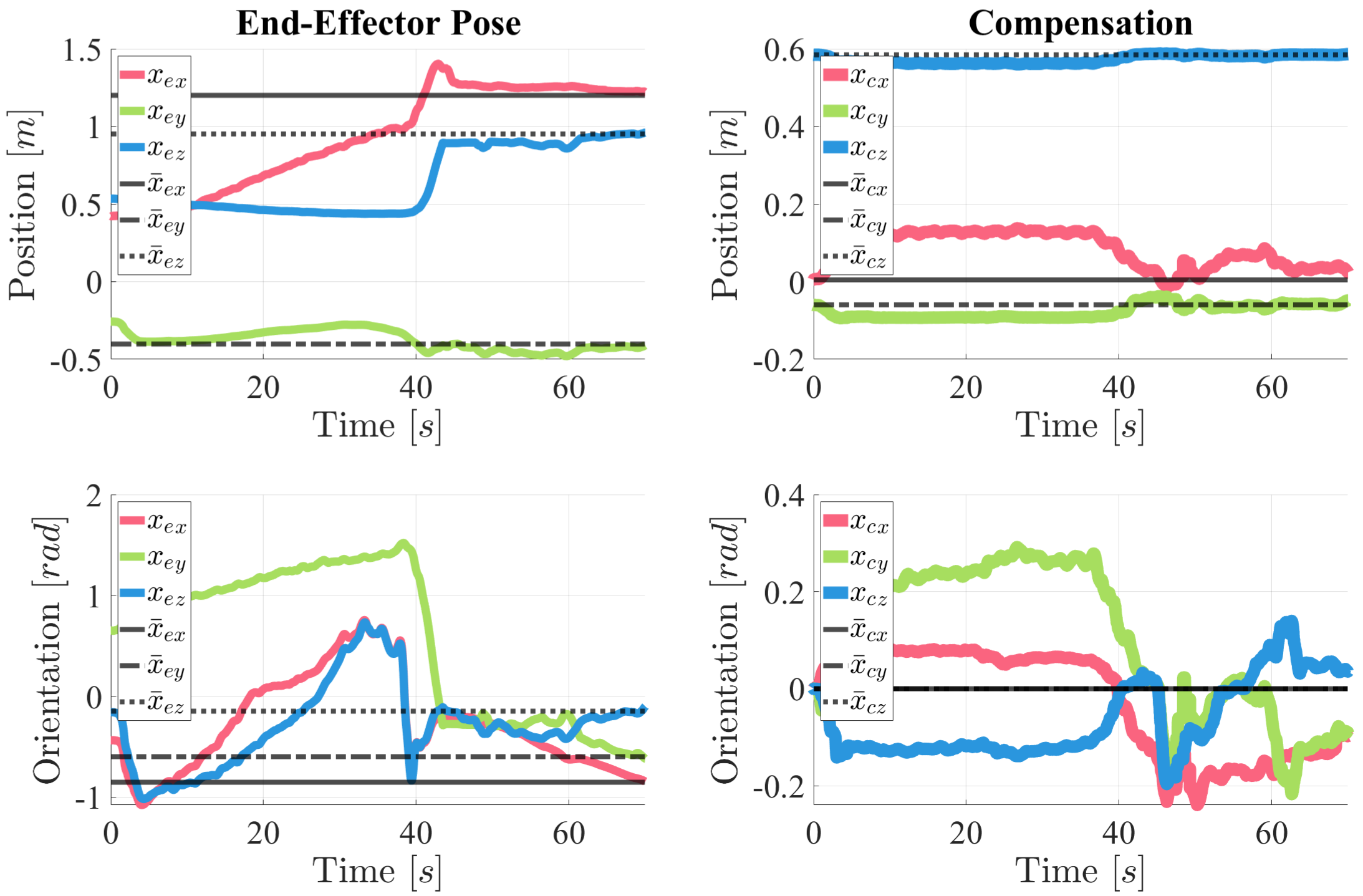}
	\caption{Actual pose of the end effector, $x_e$, and of the
	compensation frame, $x_c$, vs the desired ones, $\bar{x}_e$
	and $\bar{x}_c$, in the VR reaching task. \label{fig:vr_ego_xe}}
\end{figure}

Fig.~\ref{fig:Errors_vr_ego} shows the estimated reaching errors
$\hat{e}_e$ and the estimated compensation errors $\hat{e}_c$ for this
task.  As observed, the errors converge to zero by the end of the
task.  In  fig.~\ref{fig:vr_ego_xe}, we plot the actual pose of the end
effector and of the compensation frame (in colors) versus the desired
ones (in black).

\begin{figure}[b!]
	\centering
	\includegraphics[width = \columnwidth]{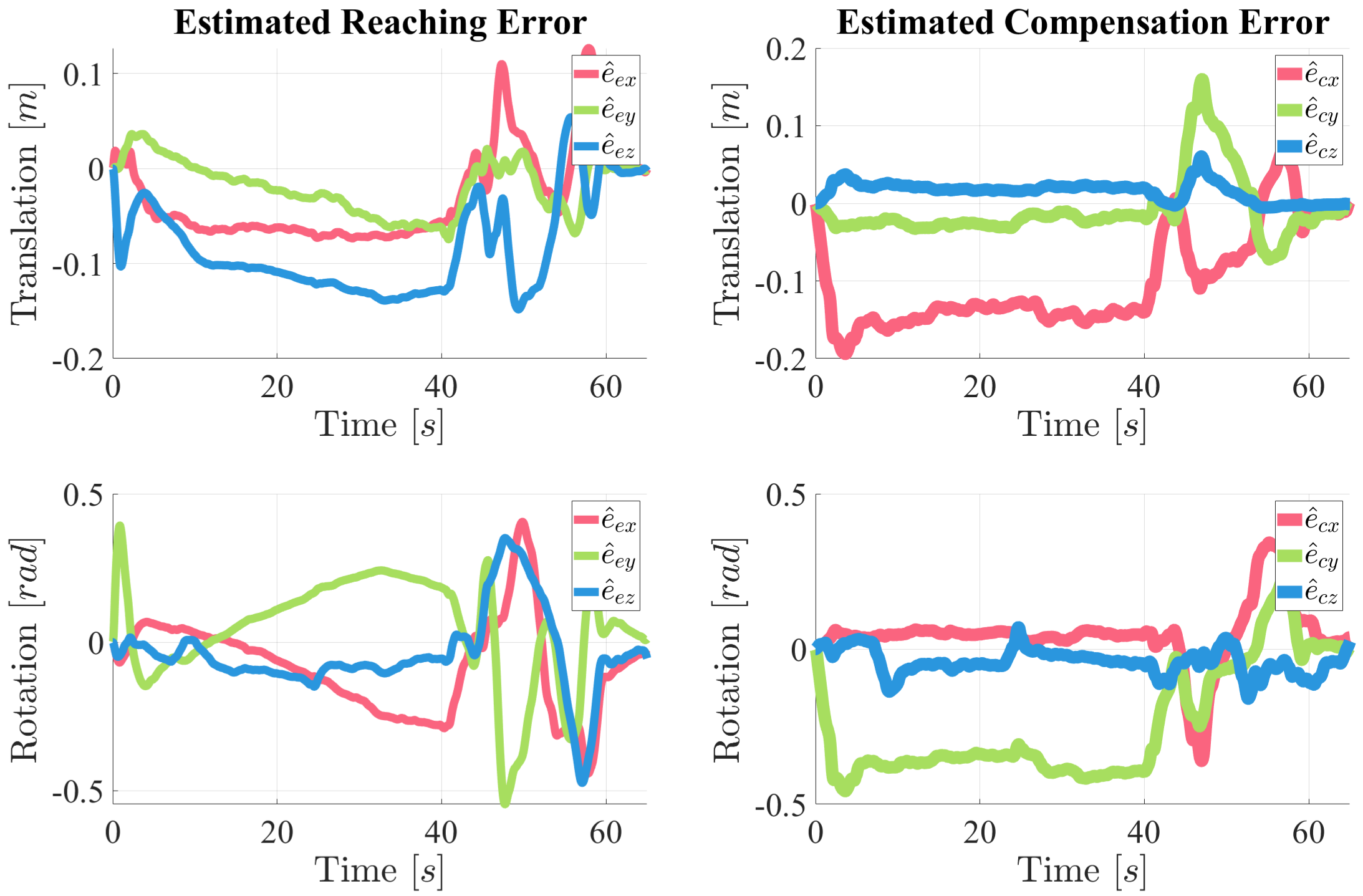}
	\caption{Estimated reaching and compensation errors for the
	real robot reaching task. \label{fig:Errors_real_ego}}

	\centering
	\includegraphics[width = \columnwidth]{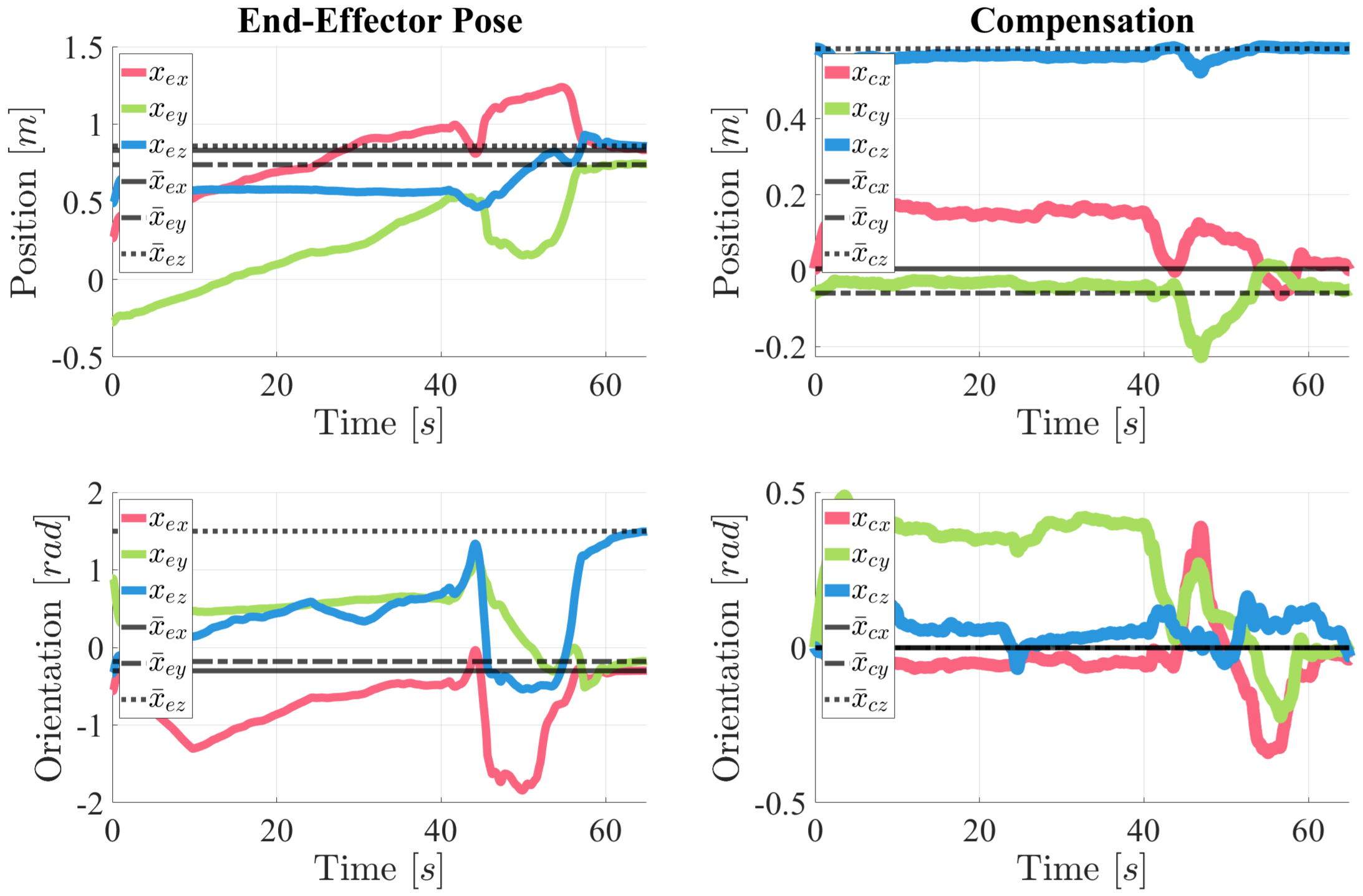}
	\caption{Actual pose of the end effector, $x_e$, and of the
	compensation frame, $x_c$, vs the desired ones, $\bar{x}_e$
	and $\bar{x}_c$ in the real robot reaching task. \label{fig:real_ego_xe}}
\end{figure}
\begin{figure*}[t!]
    \centering \includegraphics[width
	= \textwidth]{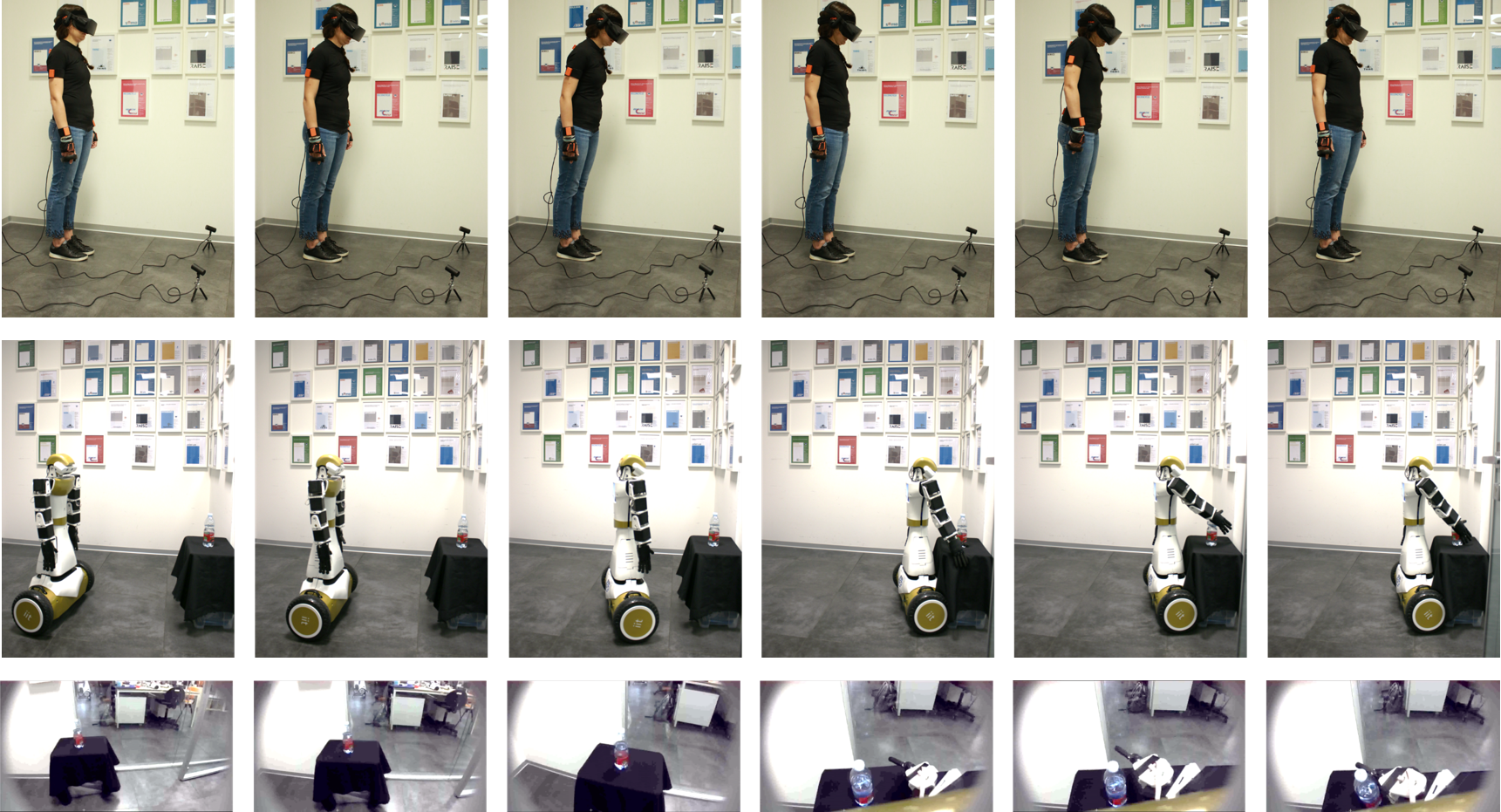}
    \caption{Six frames of
	the reaching task with the physical Alter-Ego robot. We report
	the user's motions (top), tracked via XSens, Alter-Ego's motion in
	real environment (middle), and the view from the camera placed on
	Alter-Ego's eyes (bottom). \label{fig:seq_user_ego_reach}}
\end{figure*}

Following the calibration phase and verification in simulation, we
report here results of implementation of the reaching task on a
physical version of Alter Ego.  In fig.~\ref{fig:seq_user_ego_reach},
we present a sequence of six frames illustrating the user motion (top
row) and robot motion (middle row) as the robot hand reaches the
desired goal, i.e. a water bottle on the table.  The bottom row shows
the camera view from the robot's perspective, as displayed to the user.

The estimated reaching and compensation errors are reported in
fig.~\ref{fig:Errors_real_ego}. In fig.~\ref{fig:real_ego_xe} we
present the physical hand pose $x_e$ compared to the desired
$\bar{x}_e$, as well as the actual compensation frame $x_c$ versus the
target $\bar{x}_c$. As it can be observed, the method drives the hand
to reach for the distant target using both wheels and arm movements to
accommodate for both position and orientation, using only the user's
shoulder position as input, eventually allowing the user to go back to
a relaxed posture.

\section{Discussion} 
\label{sec:Discussion} 
Our research moved from previous work \cite{Legrand:2020} showing how
users of simple prosthetic devices tend to make up for missing DoFs by
performing so-called compensatory motions with the functional parts of
their body, and how these motions can be used to drive the active
prosthetic limbs and reduce compensation. 

In this paper we extend this intuition to the control of more complex
systems where humans and robots take physical part. We outline a
theorethical framework encompassing a dynamical system where both
human and robot dynamics are integrated, design its control, and
validate it via both virtual experiments, where human motions tracked
via IMU sensors command virtual prostheses with 3, 4 and 7 DoFs,
respectively.

The generality of the approach allowed us to further extend this
algorithm to a broader range of applications, basically including all
combinations of human and robot kinematics, as depicted in
fig.~\ref{fig:startModel}. In our experiments we focused on the
control of a robot avatar, regarded as a ``whole--body prosthesis'',
using only few compensatory motions of the user.  

The functionality of the approach was demonstrated through
experiments, where the measured and estimated compensation errors
largely coincide, and the estimate of the (unmeasurable) reaching
error is reduced to zero. The controller generates robotic commands
based on user motions, and, as soon as the user returns to their
relaxed posture, the robotic joints stop moving.

Our results demonstrate that estimating the reaching error based on
user's compensatory motions is indeed possible, and that this makes it
possible to guide robotic devices to achieve the target the user aims
at, without that being ever explicited. Our approach allows the user
to focus on the desired goal, without the need to directly control
each joint indipendently, as done in conventional prosthetic
control. The user's interaction with robotic devices is thus
substantially simplified, enhancing usability and efficiency.

Given the user's crucial role in triggering the control, the
framework's effectiveness heavily depends on the user's residual
motion abilities. This implies that the control system should be
tailored on the user. Tuning weighting matrices in the controller
affects the behaviour of the human-robot integrated system, which can
be used to match certain task requirements. For instance, favoring the
use of wheel motions over arm joints can enhance performance for
specific applications where the robot should navigate the robot in a
large free space before reaching the final target.

Another crucial point is the choice of the compensatory frame. In our
experiments we placed this frame on the user's right
shoulder. However, further studies could explore alternative
positions, depending on the specific residual motions the user
controls with ease, and on the robotic system to be controlled. This
flexibility in the compensatory frame's placement could optimize the
control strategy, tailoring it more closely to the individual user's
capabilities and the requirements of different tasks.

Possible applications of the methods described here are to assist
people with disabilities that limit their motions, such as those with
neurological disorders. Preliminary tests (not described here) showed
that a potential issue in this case are involuntary reflex motions
these individuals might experience. Implementing an algorithm for
motion discrimination can significantly help in this context. The 
implementation would distinguish between intentional movements and
involuntary reflexes, ensuring that only purposeful motions trigger
the robotic commands. Additionally, integrating a deadband zone around
the user's target posture can facilitate controller deactivation,
particularly when significant residual motions are present. This
deadband zone can also help in discriminating voluntary motions from
the involuntary ones of the shoulders, associated from example with
breathing \cite{Farshchiansadegh:2014}. Further research into
incorporating kinematics studies in the control process could improve
robotic movements, enhancing overall system performance.

Lastly, it is worth to point out that the hardware platforms used to
demonstrate our method in this paper are mere examples, and other
developments could generalize and/or simplify the apparatuses - e.g.,
using visual-based motion capture could replace IMU sensors and
minimize hardware, to make the system more practical and
user-friendly.

\section{Conclusion} 
\label{sec:Conclusions}
This paper demonstrates the potential of a general control framework
for systems integrating human and robotic bodies of different shapes
and DoFs. Simulations and experiments illustrate its applicability
across different types of robots. A main characteristic of looking at
the human and robot as parts of the same system is that redundancy in
the overall systems makes it possible to reach and maintain a desired
hand target while minimizing the user's effort and discomfort.
Possible applications are various, including e.g. ergonomics of
teleoperation. Another important possible application is to assistance
and rehabilitation.

To enhance the controller's robustness, however, further tests are
necessary. These will involve conducting user case studies to evaluate
the usability and feasibility of this framework in various
applications.  Among the planned tests, we aim to work with prosthetic
users to directly assess the control's functionality and compare it
with traditional prosthetic control techniques. Ongoing studies will
also focus on assistance to patients with limited motion abilities,
including multi-subjects studies to evaluate the usability of the
Alter-Ego robot by individuals with spinal cord injuries or
patients with amyothrophic lateral sclerosis (ALS). 

\section*{Acknowledgments}
This work was supported in part by the European Research Council
Synergy Grant Natural BionicS (NBS) project (Grant Agreement
No. 810346). \\ 
This work was carried out within the framework of the project ``RAISE - Robotics
and AI for Socio-economic Empowerment'' and has been supported by European
Union - NextGenerationEU.\\
Funded by the European Union - NextGenerationEU. However, the views
and opinions expressed are those of the authors alone and do not necessarily
reflect those of the European Union or the European Commission. Neither the
European Union nor the European Commission can be held responsible for them.\\
This work was supported by the Italian Ministry of Research, under the complementary actions to the NRRP  “Fit4MedRob - Fit for Medical Robotics” Grant (\# PNC0000007). \\

Thanks to Giovanni Rosato, Manuel Barbarossa and Eleonora Sguerri for the technical support.

\bibliography{bibliography}

\end{document}